\documentclass[letterpaper]{article} 
\usepackage{aaai2026}  
\usepackage{times}  
\usepackage{helvet}  
\usepackage{courier}  
\usepackage[hyphens]{url}  
\usepackage{graphicx} 
\usepackage{natbib}  
\usepackage{caption} 
\usepackage{algorithm}
\usepackage{algorithmic}
\usepackage{graphicx}
\usepackage{marvosym}
\usepackage{subfigure}
\usepackage{subcaption}
\usepackage{inconsolata}
\usepackage{amsmath}
\DeclareMathOperator*{\argmax}{arg\,max}
\usepackage{booktabs}
\usepackage{multirow}
\usepackage{makecell}
\usepackage{microtype}
\usepackage{graphicx}
\usepackage[utf8]{inputenc}
\usepackage[most]{tcolorbox}
\newtcolorbox{mybox}[1][]{
breakable,
  arc=1mm,
  boxrule=1pt,
  colback=yellow!14,
  colframe=black!80,
  fonttitle=\bfseries,
  title=#1,
  left=1mm,
  right=1mm,
  top=1mm,
  bottom=1mm
}

\usepackage{newfloat}
\usepackage{listings}
\DeclareCaptionStyle{ruled}{labelfont=normalfont,labelsep=colon,strut=off} 
\floatstyle{ruled}
\newfloat{listing}{tb}{lst}{}
\floatname{listing}{Listing}
\title{MIDB: Multilingual Instruction Data Booster for Enhancing Cultural Equality in Multilingual Instruction Synthesis}

\author{
    Yilun Liu\textsuperscript{\rm 1}\thanks{Equal contribution.},
    Chunguang Zhao\textsuperscript{\rm 1$*$},
    Xinhua Yang\textsuperscript{\rm 1$*$},
    Hongyong Zeng\textsuperscript{\rm 1},
    Shimin Tao\textsuperscript{\rm 1}, 
    Weibin Meng\textsuperscript{\rm 1},\\
    Minggui He\textsuperscript{\rm 1},
    Yan Yu\textsuperscript{\rm 1},
    Hongxia Ma\textsuperscript{\rm 1},
    Li Zhang\textsuperscript{\rm 1},
    Daimeng Wei\textsuperscript{\rm 1},
    Boxing Chen\textsuperscript{\rm 2}
}

\affiliations{
    \textsuperscript{\rm 1}Huawei, China
\\ \textsuperscript{\rm 2}Huawei Canada, Canada\\
\texttt{\{liuyilun3, zhaochunguang4, yangxinhua2\}@huawei.com}
}

\begin{document}
\maketitle

\begin{abstract}
Despite doubts on data quality, instruction synthesis has been widely applied into instruction tuning (IT) of LLMs as an economic and rapid alternative. Recent endeavors focus on improving data quality for synthesized instruction pairs in English and have facilitated IT of English-centric LLMs. However, data quality issues in multilingual synthesized instruction pairs are even more severe, since the common synthesizing practice is to translate English synthesized data into other languages using machine translation (MT). Besides the known content errors in these English synthesized data, multilingual synthesized instruction data are further exposed to defects introduced by MT and face insufficient localization of the target languages, leading to cultural inequality in trained LLMs. In this paper, we propose MIDB, a Multilingual Instruction Data Booster to automatically address the quality issues in multilingual synthesized data. MIDB is trained on around 36.8k revision examples across 16 languages by human linguistic experts, thereby can boost the low-quality data by addressing content errors and MT defects, and improving localization in these synthesized data. Both automatic and human evaluation indicate that not only MIDB steadily improved instruction data quality in 16 languages, but also the instruction-following and cultural-understanding abilities of multilingual LLMs fine-tuned on MIDB-boosted data were significantly enhanced, suggesting an improved linguistic and cultural equality.

\end{abstract}

\begin{links}
    \link{Code \& Datasets}{https://github.com/zhaocorey/MIDB}
\end{links}

\section{Introduction}
\label{sec:introduction}

Large language models (LLMs) have made significant strides in their performance in English, achieving impressive capabilities across a range of natural language processing (NLP) tasks~\cite{achiam2023gpt,deepseek2025r1}. However, the multilingual abilities of most LLMs remain relatively underdeveloped~\cite{lai2024llms}, particularly due to the predominance of English in the pretraining data used by many popular open-source LLMs, such as the LLaMA series~\cite{touvron2023llama}. For example, in the case of LLaMA-2~\cite{touvron2023llama2}, the ratio of non-English languages in its pretraining corpus is merely around 2\%, which could significantly limit the abilities of downstream models to handle requests from diverse linguistic communities, and potentially leads to linguistic biases and insufficient cultural representations~\cite{rystrom2025multilingual}.

\begin{figure}
    \centering
    \includegraphics[width=1\linewidth]{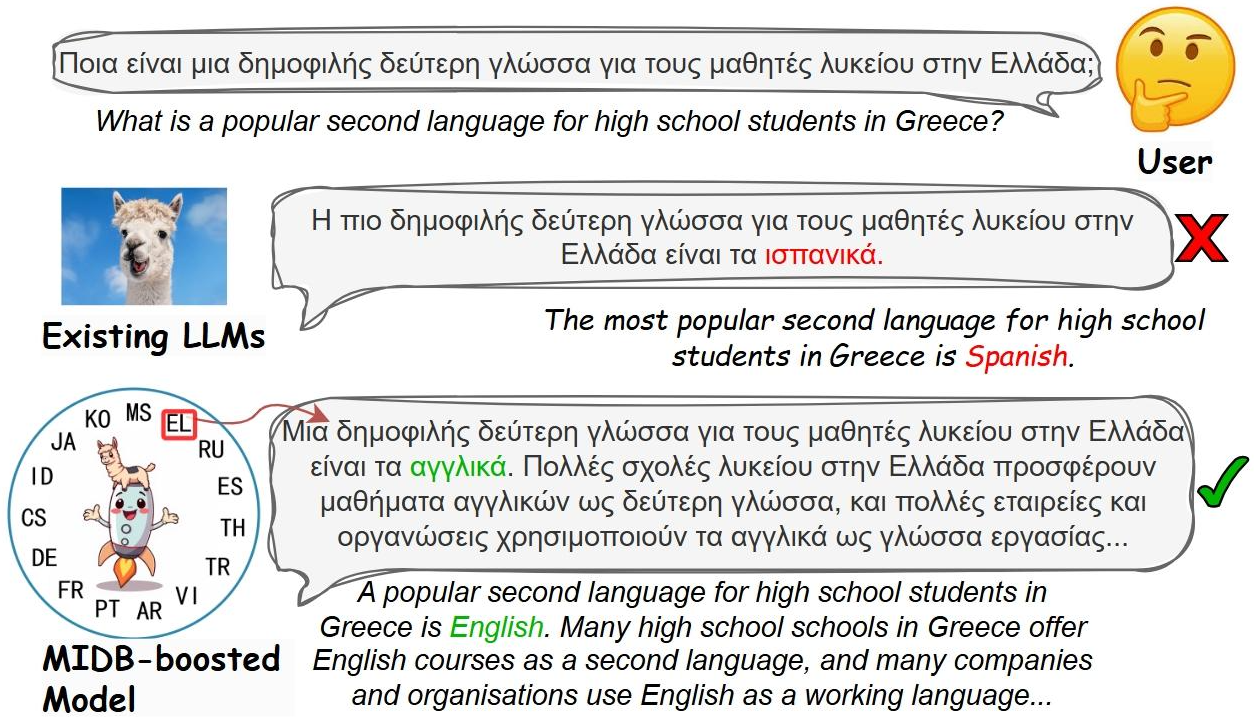}
    \caption{An example suggesting improved cultural equality in MIDB-boosted LLMs: it successfully identified a popular second-language in the cultural context of Greece. EL, RU, ES, \emph{etc.}, are language codes. See code-name mapping of the 16 supported languages in Appendix~\ref{language_dict}.}
    \label{fig:cultrual case}
\end{figure}

A widely-adopted mitigating approach is multilingual instruction tuning (IT), which introduces multilingual instruction pairs (\emph{i.e.}, instruction-response pairs) during the post-training phase to enhance the model's ability to understand and process multiple languages~\cite{chen2024monolingual,shaham2024multilingual}. However, obtaining high-quality multilingual instruction data remains a significant challenge, especially for low-resource languages due to limited linguistic resources. Compared to monolingual contexts (\emph{e.g.}, English), manually creating high-quality instruction datasets for multiple languages naturally incurs higher social utility costs and resource demands~\cite{ustun2024aya}. In contrast, using instruction synthesis techniques to automatically generate large-scale multilingual instruction data presents a more cost-effective and scalable alternative.

While research on instruction synthesis for English has flourished~\cite{wang-etal-2023-self-instruct, alpaca, xu2023wizardlm}, multilingual instruction synthesis has largely relied on machine translation (MT) to adapt synthesized English datasets to target languages~\cite{lai2024llms, chen2024monolingual, huo2025enhancing}. A common approach involves translating the English Alpaca dataset~\cite{alpaca}, which contains 52k instruction pairs generated by LLMs, into other languages. Although MT provides a practical solution for rapid adaptation, it introduces data quality issues such as MT-introduced defects and insufficient cultural localization (as they tend to reflect cultures represented by English). As noted in prior studies~\cite{zhou2023lima,li2024quantity,liu2024coachlm}, such data quality issues could significantly reduce the effectiveness of multilingual IT. Fully resolving these issues may require extensive human involvement, which is impractical for large scales—particularly in multilingual contexts.

To address this, we introduce MIDB, an automatic Multilingual Instruction Data Booster aimed at improving the quality of synthesized instruction data for multilingual IT. Inspired by data engineering techniques in English~\cite{liu2024coachlm,ge2024clustering}, where LLMs learn from human ratings or revision patterns, we collaborated with linguistic experts to build a dataset of 36.8k manual revision examples across 16 languages, including low-resource ones. These revisions target low-quality instruction pairs, correcting content issues (\emph{e.g.}, accuracy, richness, relevance) and MT-induced defects (\emph{e.g.}, fluency, correctness). These examples enable LLMs to learn human boosting strategies and enhance multilingual instruction data automatically. We also ensured that the dataset preserves language-specific features and emphasizes cultural and linguistic localization. This dataset is denoted as the Multilingual Expert Boosted (MEB) dataset.

We then used this MEB dataset to train MIDB to automate the enhancement of synthesized data quality for multilingual IT, leading to significant improvements on both multilingual data quality and model performance on multilingual and cultural-understanding abilities. In addition, since existing multilingual evaluation benchmarks for instruction-following abilities of LLMs also suffer from defects introduced by MT, as is observed by \citet{chen2024good}, we conducted manual localization of two most popular benchmarks, AlpacaEval~\cite{dubois2023alpacafarm} and MT-Bench~\cite{zheng2023judging}, into 16 languages with the help of professional translators.

\section{Social Impact of MIDB}
\label{sec:socialImpact}
MIDB addresses critical linguistic and cultural inequalities in multilingual AI systems, directly aligning with global priorities in social welfare and underserved communities. By tackling flaws in multilingual instruction synthesis, MIDB demonstrates potentials in two aspects:

(1) \textbf{Bridge the English-dominated Digital Divide.} As highlighted by a report by Stanford~\cite{stanford2025digitaldivide}, The non-English speakers in the world may suffer a digital divide caused by English-dominated LLMs. “The ChatGPTs and Geminis of the world work well for the 1.52 billion people who speak English, but they underperform for the world’s 97 million Vietnamese speakers”, the report states. The impact for non-English communities may be beyond inconvenience, but a systematic exclusion from the gains by AI revolution, such as economic, educational and clinical opportunities, compared with those who are fluent in English. MIDB’s universal improvements on multilingual performance and released assets provide a cost-effective solution to this exclusion, enabling more localized AI applications (\emph{e.g.}, LLM's Performance in Vietnamese increased by 25.9\% with MIDB).

(2) \textbf{Mitigating Cultural Inequality.} Current machine-translated instruction data (\emph{e.g.}, from English) often erases cultural nuances, or even leads to cultural biases (\emph{e.g.}, US-centric~\cite{rystrom2025multilingual}). As shown in Fig.~\ref{fig:cultrual case}, Alpaca, an existing LLM, states that a popular second language in Greece is Spanish. This mistake may come from its English-originated training data, where Spanish is indeed a popular second language in the US. Such cultural inequalities may “collapsing cultural diversity into one big blob”~\cite{stanford2025digitaldivide}, which is especially unsettling when LLMs are applied in high-stake scenarios like education. In contrast, MIDB’s focuses on cultural localization with human experts may help enhance cultural diversity in multilingual training data (\emph{e.g.}, in Section~\ref{sec: culture evaluation}, MIDB-boosted LLMs' accuracy for cultural questions increased by 19.5\%  for five non-English cultures).

\section{Related Work}
\label{sec:relatedWork}

\subsection{Multilingual Instruction Tuning}

To improve the multilingual abilities of existing foundation LLMs, various training methods for enhancing multilingual IT are proposed. \citet{chen2024monolingual} combined multiple MT versions of the Alpaca dataset into training one multilingual model and achieved improved performance compared with monolingual baselines. Upon this combination, \citet{zhu2023extrapolating} further involved translation instructions into the training set, requesting the model to translate source sentences in English to sentences in target languages, to help transfer the knowledge learned in English. To make this transferring process more explicit, \citet{zhang2024plug} composed specialized instruction pairs that ask the model to first process instructions in a pivot language (\emph{e.g.}, English) and then produce response in the target language.

Despite improved training recipes, most existing methods directly adopt the synthesized Alpaca dataset and its MT versions as the training set, hampering effectiveness of their approaches due to known data quality issue. Our work fills this blank by focusing on data quality issue multilingual IT.

\subsection{Instruction Data Synthesis}

The technique of instruction data synthesis facilitates training of LLMs by generating instruction pairs using powerful LLMs, saving labors of human annotation. The pioneering attempt of instruction synthesis is Self-Instruct~\cite{wang-etal-2023-self-instruct}, leveraging LLMs to produce instruction pairs given a small manually-written seed dataset. Subsequently, the Alpaca project~\cite{alpaca} utilized the Self-Instruct strategy to generate 52k instruction pairs using LLMs. A lot of methods then focus on improving this pipeline, either by adding quality-filtering modules~\cite{chen2023alpagasus,ge2024clustering} or increasing generation diversity~\cite{xu2023wizardlm,liu2024coachlm}. However, most existing technologies for instruction synthesis are targeted for the language of English, leaving the synthesis of multilingual instruction data underexplored. Compared with these existing methods, our work focuses on high-quality instruction synthesis for multiple languages.

We also noticed the concurrent work of \citet{feng2025culfit}, who proposed a framework to synthesize cultural-related instruction pairs with critiques and translate them into multilingual data using MT. Our work, compared with theirs, works closely with language experts to solve the fine-grained problems such as fixing local expressions and cultural mismatching terms, potentially mitigating hallucinations by MT.

\begin{figure}[tb]
    \centering
    \subfigbottomskip=-2pt 
    \subfigcapskip=0pt 
    \subfigure[\textbf{MIDB model training}]{
    \includegraphics[width=0.7\linewidth]{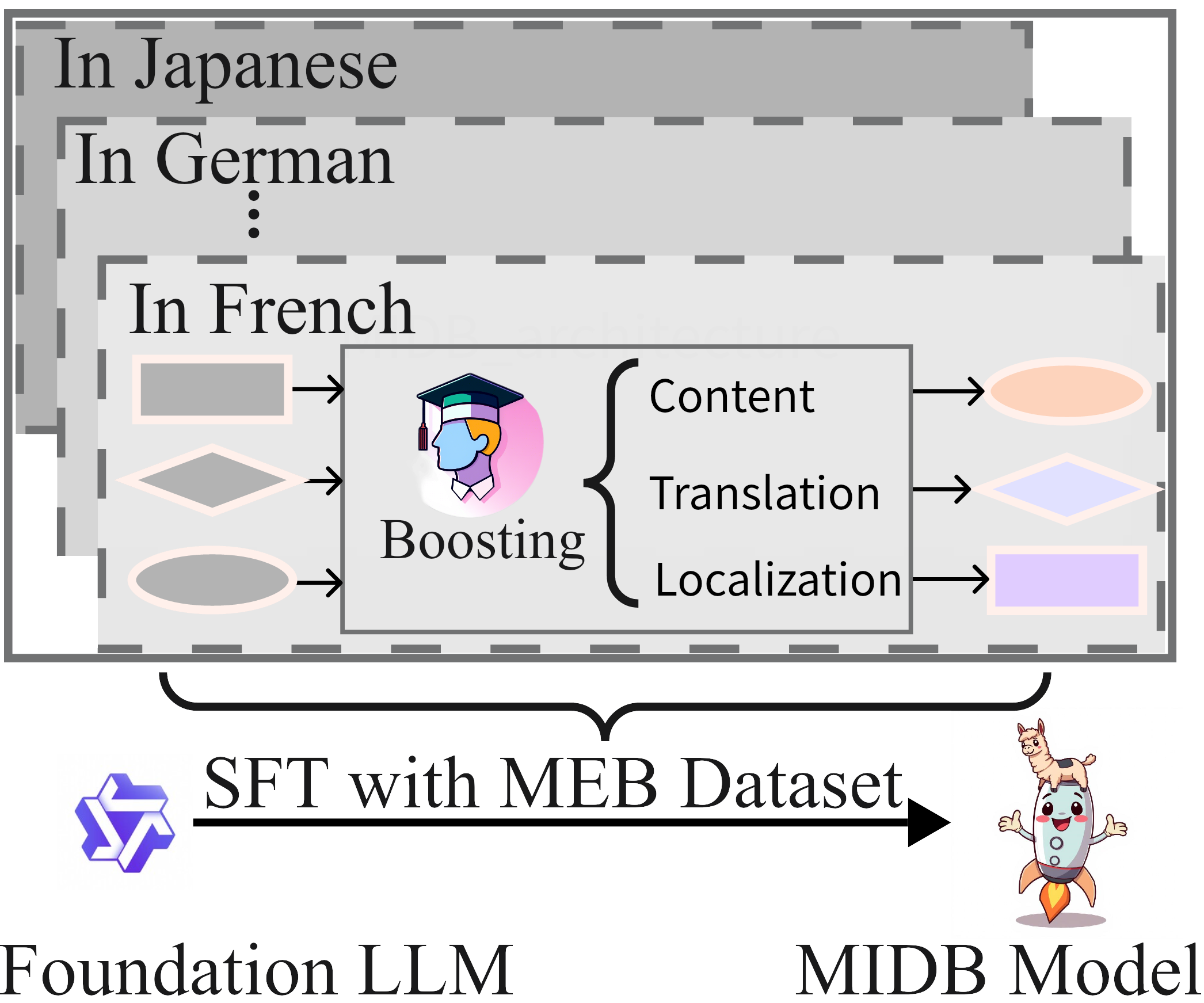}}\\
    \subfigure[\textbf{MIDB model applied in data boosting}]{
    \includegraphics[width=0.9\linewidth]{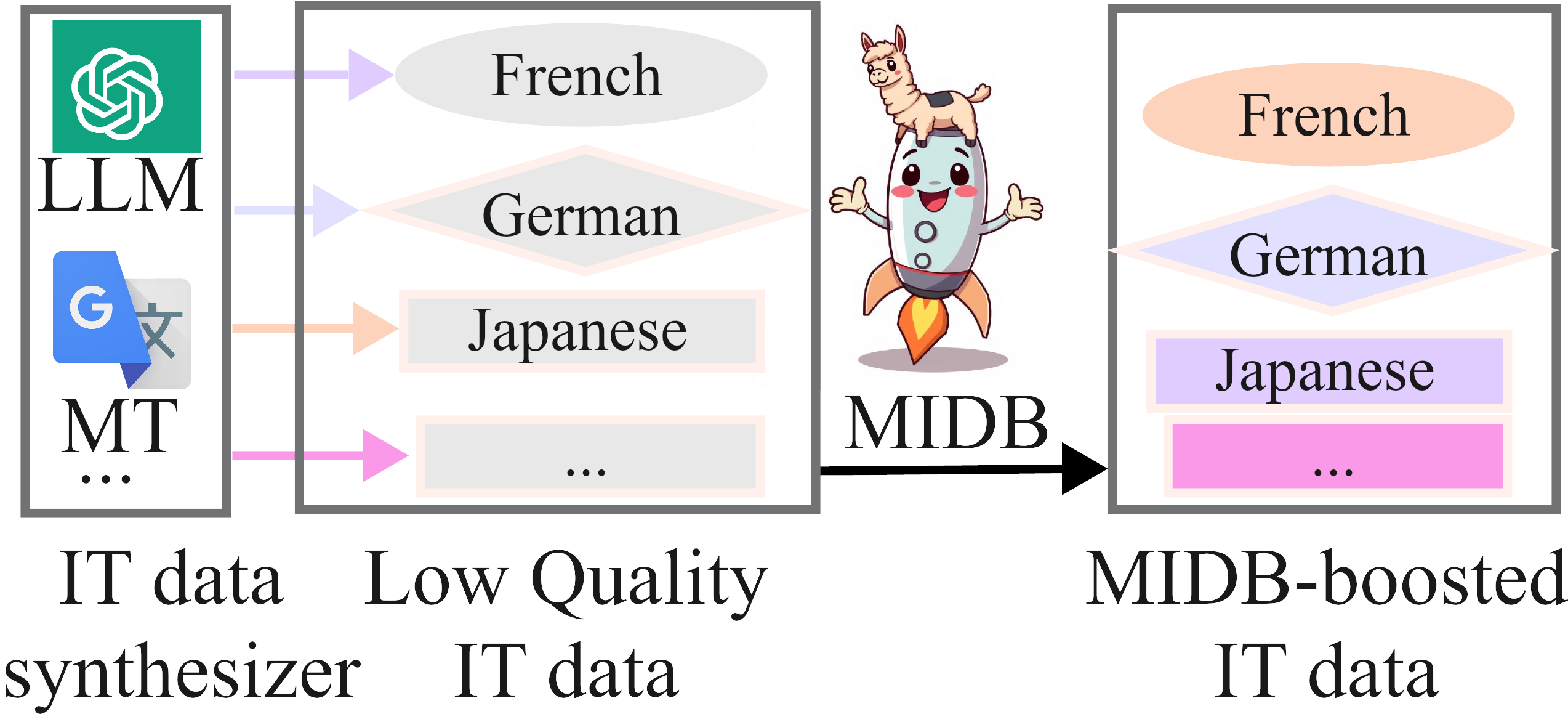}}
    \caption{Illustrations on (a) training stage and (b) inference stage of MIDB.}
    \label{fig:MIDB_architecture}
\end{figure}

\section{Methodology}
\label{sec:method}

The overview of the training and inference of MIDB is shown in Fig.~\ref{fig:MIDB_architecture}. To construct training dataset for MIDB, a thorough manual correction and revision on subsets of the machine-translated Alpaca datasets (from English) was firstly conducted (Section~\ref{sec:expert_profile}-\ref{sec:training_set_construction}). Section~\ref{sec:training_of_MIDB} discusses the training of MIDB. And Section~\ref{sec:construction_of_testset} discusses the construction of our multilingual benchmarks.

\subsection{Profile of Involved Multilingual Human Experts and Task Allocation Strategy}\label{sec:expert_profile}
To accomplish the construction of the MEB dataset and two benchmarks, and human evaluation, we recruited and actively worked with a group of language experts (possessing an average of 6.5+ years' experience), coming from an international corporation's language service center with expertise in translation, localization, and editing. Experts were strategically allocated to the three tasks based on their proficiency while maintaining language coverage and avoiding bias through strict task separation. Quality was assured via a two-round review-rebuttal process and third-party oversight during annotation. Appendix~\ref{sec:detail_expert_profile} discusses full details of expertise, allocation strategies and quality control.

\begin{figure}[tb]
     \subfigbottomskip=-2pt 
     \subfigcapskip=-2pt 
    \centering
    \includegraphics[width=0.85\linewidth]{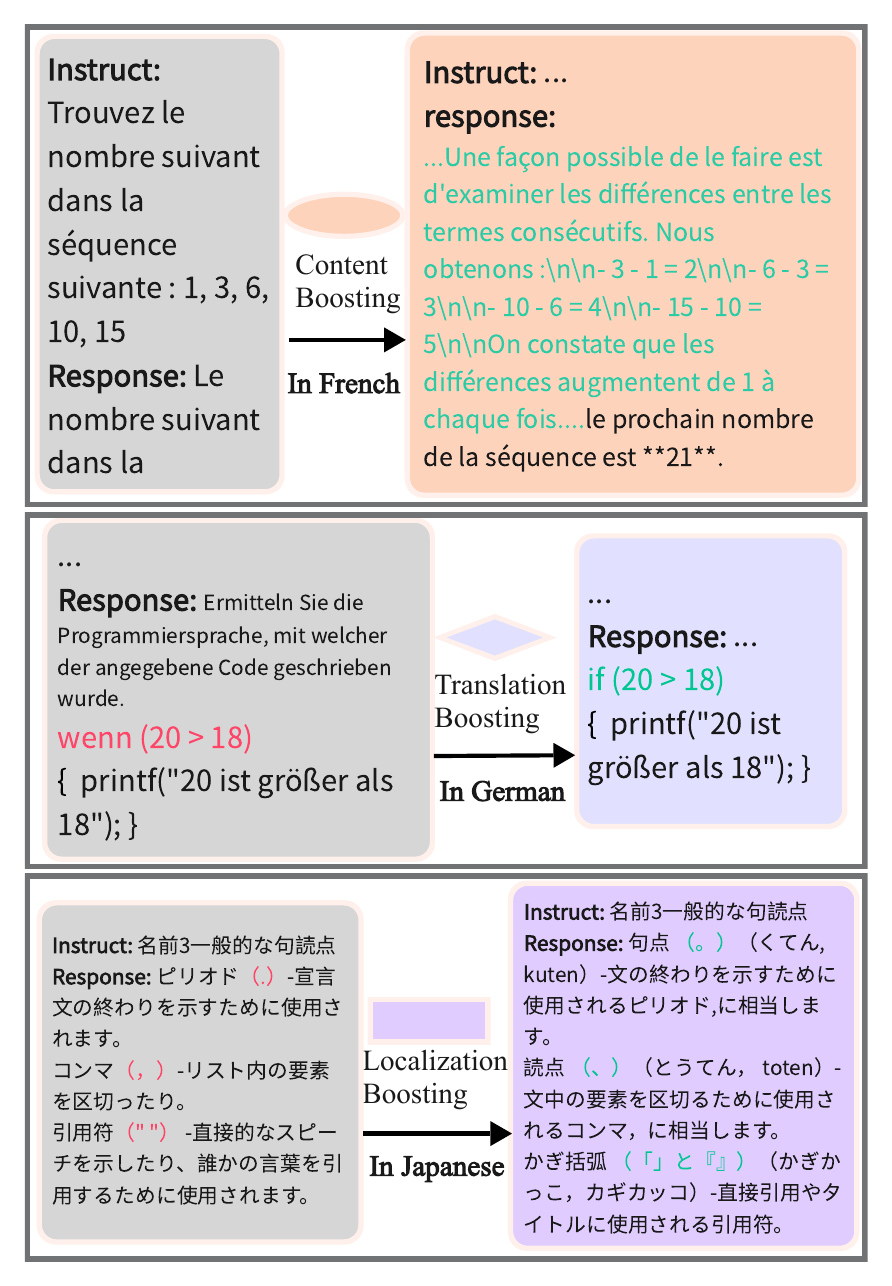}
    \caption{Typical issues addressed in MEB Dataset.}
    \label{fig:data_examples}
\end{figure}

\subsection{Building MEB Dataset}\label{sec:training_set_construction}

\paragraph{Preliminary Data Quality Study with Experts}

As the initial step, our language experts inspected several popular multilingual IT datasets, which were machine-translated from the synthesized English Alpaca Dataset~\cite{alpaca}. We then discussed with experts and identified several notable data quality issues within these samples and concluded them into three most common categories (Fig.~\ref{fig:data_examples} illustrates typical cases from each category, along with expert suggestions):

\textbf{(1) Content errors in the source (\emph{e.g.}, English) datasets.} Synthesized instruction datasets like Alpaca often contain content errors and defects due to LLM hallucinations during data generation~\cite{chen2023alpagasus, liu2024coachlm, ge2024clustering}. These errors range from surface-level issues like formatting and grammatical mistakes~\cite{AlpacaDataCleaned}, to deeper defects such as logical inconsistencies, factual inaccuracies, and one-sided explanations~\cite{liu2024coachlm}. While some of these can be addressed with rule-based filtering, deeper issues still require human intervention. The first example in Fig.~\ref{fig:data_examples} illustrates a typical deeper defects, lack of comprehensiveness due to absence of intermediate process.

\textbf{(2) Defects introduced by MT.} Despite advances, current commercial MT systems or off-the-shelf LLMs still have significant defects in translation. For example, \citet{lai2024llms} found that multilingual Alpaca datasets translated via Google Translate API had an average error rate of around 30\% in five low-resource languages. Such high translation error rates can lead to cascading issues. The second example in Fig.~\ref{fig:data_examples} addresses a common MT error, where the conditional statement "if" were mistakenly translated into the target language, potentially leading to catastrophic errors in the code compiler.

\textbf{(3) Insufficient Localization.} Direct translation of instruction data often leads to inadequate localization, as source instruction pairs tend to reflect the cultural and knowledge contexts of the source language (\emph{e.g.}, English), which can lead to mismatches in the target language context. In the third case in Fig.~\ref{fig:data_examples}, the response of a Japanese instruction, "List 3 common punctuation marks", mistakenly included English punctuation marks. With the help of language experts, the answer was accurately localized to incorporate Japanese-specific punctuation marks.

\begin{table}[h]
\centering
\resizebox{\linewidth}{!} {%
\begin{tabular}{c | c | c c}
\toprule
\textbf{Category} & \textbf{Ratio} & \multicolumn{2}{c}{\textbf{Criteria}}  \\
\midrule
\multirow{5}{*}{\makecell{Content \\ Boosting}} & \multirow{5}{*}{\makecell{22.9\%}}  
 & Contextualization & Relevance           \\ 
& & Feasibility & Timelineness              \\ 
& & Humanization & Comprehensiveness        \\
& & Richness  & Correctness                 \\ 
& & Readability & Safety                    \\ 
 \midrule
\multirow{3}{*}{\makecell{Translation \\ Boosting}} & \multirow{3}{*}{\makecell{24.4\%}} 
 & Fluency & Grammar                                  \\ 
& & Translation Elegancy & Omitted translation     \\
& & Spelling & Incorrect translation           \\
 \midrule
\multirow{2}{*}{\makecell{Localization}} &  \multirow{2}{*}{\makecell{52.7\%}} 
 & Culture localization & Geocultural term repair          \\ 
& & Ideology localization & Expression localization      \\
\bottomrule
\end{tabular}
}
\caption{Manual evaluation and enhancement criteria for quality issues in multilingual synthesized datasets. }
\label{tab:MEDB_criteria}
\end{table}

\paragraph{Criteria for Building MEB Dataset}\label{data_Criteria}
The categories derived from the typical issues above have been further summarized and listed as criteria in Table \ref{tab:MEDB_criteria} for manual enhancement: 

\textbf{The “Content Boosting” category} is primarily inspired by the criteria proposed by \citet{liu2024coachlm}, who introduced content revision criteria for English IT data and demonstrated their effectiveness through various experiments. Their English-based standards for content errors in synthesized data remain applicable in multilingual contexts. For example, criteria such as Relevance and Comprehensiveness are universally relevant and independent of the specific language. As such, we have inherited these criteria from their work.

\textbf{The “Translation Boosting” category} is derived from professional translation standards from the cooperated language service center, reflecting challenges for MT models such as "Omitted Translation" and "Translation Elegancy".

\textbf{Localization-related criteria} are the most challenging aspect of our work, mainly due to the limited availability of low-resource language experts and the lack of public instruction data. To address this, we propose four novel data-boosting criteria related to localization:

(1) \textit{Cultural Relevance}: Adapting instruction pairs to reflect local culture, including references to local music, movies, and food where appropriate.

(2) \textit{Geo-cultural Terms}: Recognizing that some entities are known by different names across regions. For instance, the Himalayas are referred to as Mount Everest in some languages. Instruction pairs should use terminology that aligns with these regional variations.

(3) \textit{Ideological Localization}: This criterion addresses differences in religion, history, and local media, where the same input may yield different responses based on these factors. Some instruction pairs may need to be entirely restructured to align with these differences.

(4) \textit{Local Expression}: Emphasizing the use of local, culturally relevant expressions instead of direct translations. These expressions help retain the unique features of each language, much like an ethnic costume showcases cultural identity.

\paragraph{Manual Enhancement Results}
A total of 16 languages were selected for inclusion in our study based on an evaluation of factors such as geographic coverage, popularity, impact, and available human resources. Notably, the selection includes four low-resource languages, ensuring coverage of all geographical regions (more details in Appendix~\ref{sec:language_distribution}).

After dedicating over 485 person-days to the construction of the MEB dataset, we curated 36.8k manually boosted instruction pairs across the 16 languages, averaging approximately 2.3k pairs per language. As shown in Table~\ref{tab:MEDB_criteria}, over 52.7\% of the instruction pairs have been localized according to the proposed criteria, including both expression localization and cultural adaptation. And 22.9\% and 24.4\% of the manual enhancements address content issues and MT defects.

\subsection{Training Design of MIDB} \label{sec:training_of_MIDB}

\paragraph{Building Training Samples for MIDB} 

\begin{figure}[htb]
    \centering
    \includegraphics[width=0.9\linewidth]{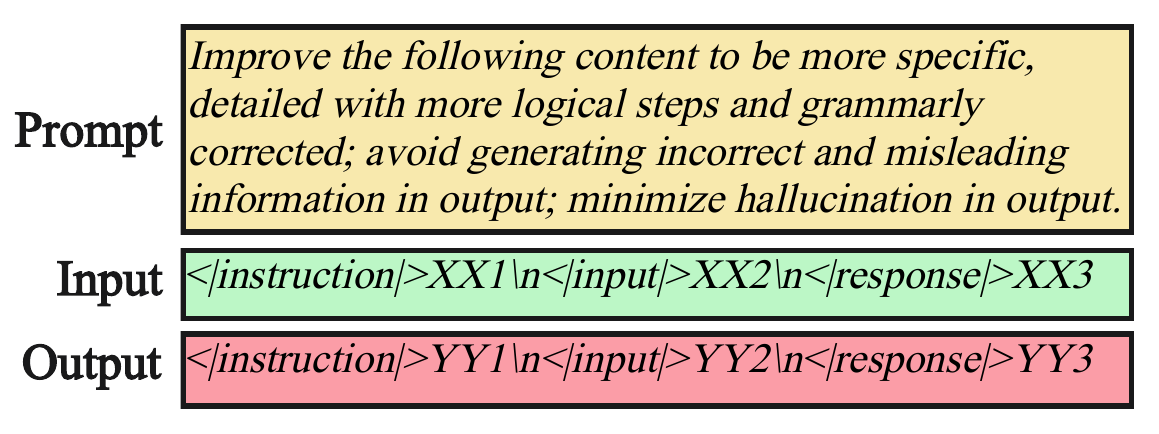}
    \caption{Template of MIDB's training samples.}
    \label{fig:MEB_dataset_template}
\end{figure}

The manual enhancement examples are subsequently transformed into training samples for MIDB. As shown in Fig.~\ref{fig:MEB_dataset_template}, a training sample consists of \textit{Prompt}, \textit{Input} and \textit{Output}. The \textit{Prompt} is adapted from \citet{liu2024coachlm}, serving as a straightforward instruction for content refinement during the training of MIDB. The \textit{Input} consists of an original instruction pair in the MEB Dataset, with its $<|$instruction$|>$, $<|$input$|>$, and $<|$response$|>$ concatenated into a string. The \textit{Output} is the corresponding manually boosted instruction pair, serving as a learning target of MIDB to activate expert-aligned multilingual instruction boosting capabilities of the foundation LLM.

\paragraph{Joint Training Goal for 16 Languages} \label{sec:training goal}
To reduce deployment cost and facilitate connections between languages, we trained MIDB as a unified model capable of boosting instruction pairs from all 16 languages. From the backbone $\theta$, the joint training goal shown in Eq.~\eqref{eq_loss} optimize it into $\theta_m$, \emph{i.e.}, MIDB. \(C_i\) represents the subset of the training samples from the \(i_\textit{th}\) language within these 16 languages. For the \(j_\textit{th}\) training sample in \(C_i\), \(x_j\) is constructed by concatenating \textit{Prompt} and \textit{Input}, and \(y_j\) is directly the corresponding \textit{Output}. Following \citet{liu2024coachlm}, only high-quality subsets are utilized in training MIDB with a quality-control coefficient $\alpha$ (defined as the top $\alpha$-percent of samples that received the most revisions). See a sensitivity analysis on $\alpha$ and backbone $\theta$ in Section~\ref{sec:albation_exp}.

\begin{equation}\label{eq_loss}
   \theta_m = \argmax_{\theta} \sum\limits_{i\in [1,16]} \sum\limits_{x_j\in C_i} \log P(y_j\,|\,x_j;\theta)
\end{equation}

\subsection{Construction of Multilingual Test Sets} \label{sec:construction_of_testset}

The evaluation encompasses three public benchmarks assessing instruction-following, multi-turn dialogue, and culture-specific understanding ability: AlpacaEval~\cite{alpaca_eval}, MT-Bench~\cite{zheng2023judging}, and BLEnD~\cite{myung2024blend}. AlpacaEval is a benchmark validated by 20k+ human judgments, aiming to assess the instruction-following ability of LLMs on general topics. MT-Bench includes 80 multi-turn questions across eight intent categories (\emph{e.g.}, coding, reasoning, knowledge), designed to challenge strong models. AlpacaEval and MT-Bench were originally designed for English without multilingual support. We acquire the open-source English questions from AlpacaEval and MT-Bench, and a team of 20 professional experts (as discussed in Section~\ref{sec:expert_profile}) spent 175 person-days translating them to 16 languages. And hence we rename them to AlpacaEval-16L and MT-Bench-16L, respectively. Notably, the BLEnD benchmark already covers five languages within the scope of our evaluation.

\section{Experiment}
\label{sec:exp}

Implementation details of MIDB are introduced in Section~\ref{sec:implementation details}. To ensure a comprehensive evaluation (setups discussed in Section~\ref{sec:Experiment Setting}), both data quality of MIDB-boosted datasets and performances of subsequently tuned LLMs are evaluated, encompassing both automatic (Section~\ref{sec:GPT Evaluation}) and manual (Section~\ref{sec:Human Evaluation}) evaluations. Section~\ref{sec: culture evaluation} examines the enhancement on cultural understanding brought by MIDB.  

Section~\ref{sec:Effectiveness on out-of-distribution dataset} verifies MIDB on an out-of-distribution dataset. In Section~\ref{sec:albation_exp}, we conduct an sensitivity analysis on critical parameters of MIDB.

\subsection{Implementation Details}\label{sec:implementation details}

We explored different backbone models $\theta$ and different quality-control coefficient $\alpha$ for MIDB. In our main implementation of MIDB, we used LLaMA3.1-8B-Instruct~\cite{grattafiori2024llama3herdmodels} as the backbone model, which has 8B parameters and possess strong multilingual abilities, and set quality-control coefficient to 30\%.
To efficiently adapt the backbone LLMs, we employed LoRA~\cite{hu2021loralowrankadaptationlarge}, a parameter efficient fine-tuning technique, with a rank of 64. MIDB was trained for three epochs with a learning rate of $4\times10^{-4}$ and global batch size is set to 128. For training the instruction-following models, we utilized the same settings as the official Alpaca repository~\cite{alpaca}, with the exception of using different instruction datasets and using LLaMA3-8B as the backbone. The multilingual Alpaca datasets and Dolly datasets used during evaluation was translated by advanced LLMs from English. And a beam size of one was used for decoding across all models.

\begin{figure*}[htbp]
    \centering
    \includegraphics[trim=10mm 4mm 10mm 10mm, clip, width=\linewidth]{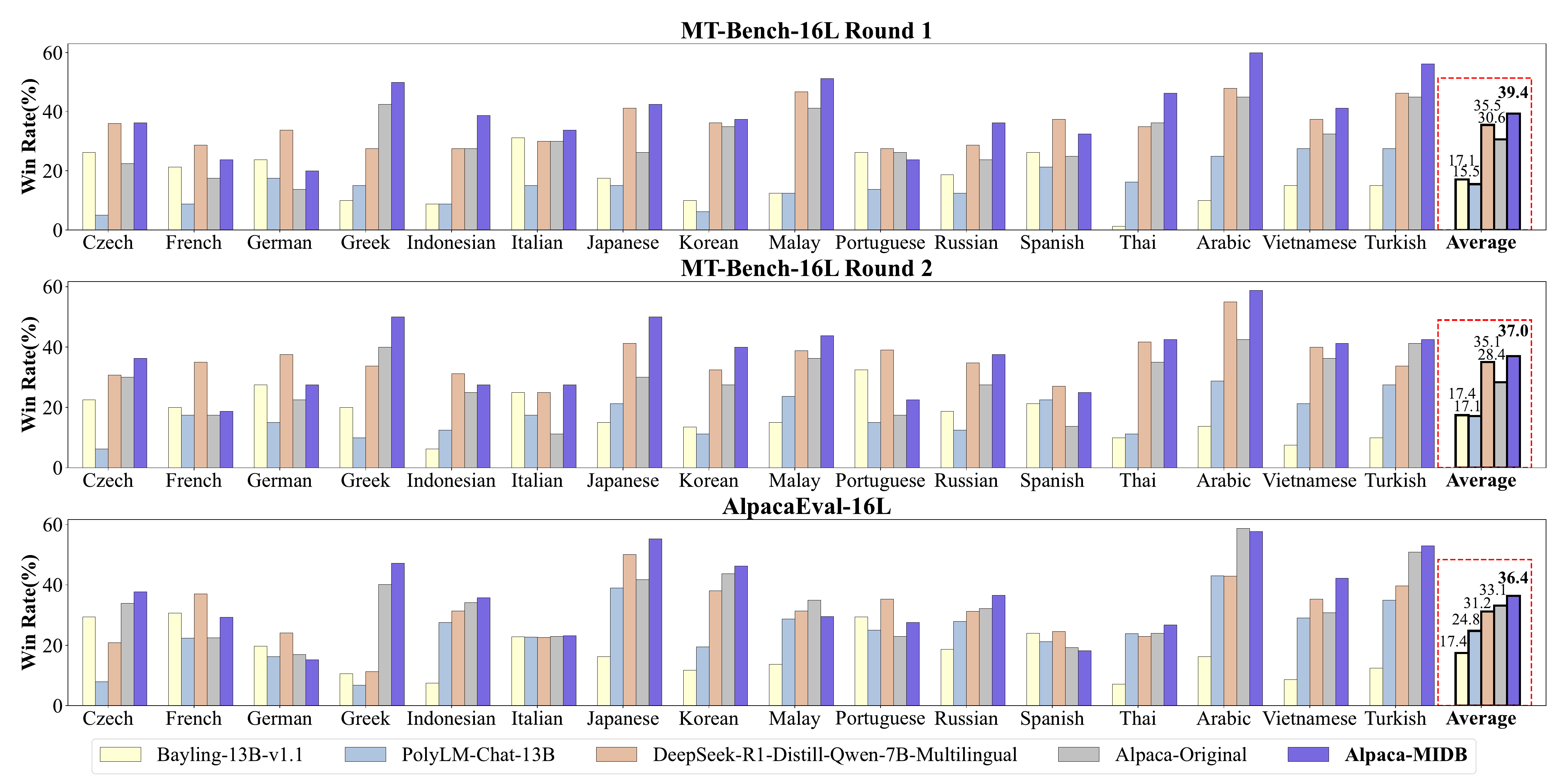}
    \caption{LLM-as-judge's evaluation on LLMs trained with MIDB-boosted/pre-boosted Alpaca datasets and 3 strong LLMs.}
    \label{fig:model ability}
\end{figure*}

\begin{figure}[tbp]
    \centering
    \includegraphics[width=0.9\linewidth]{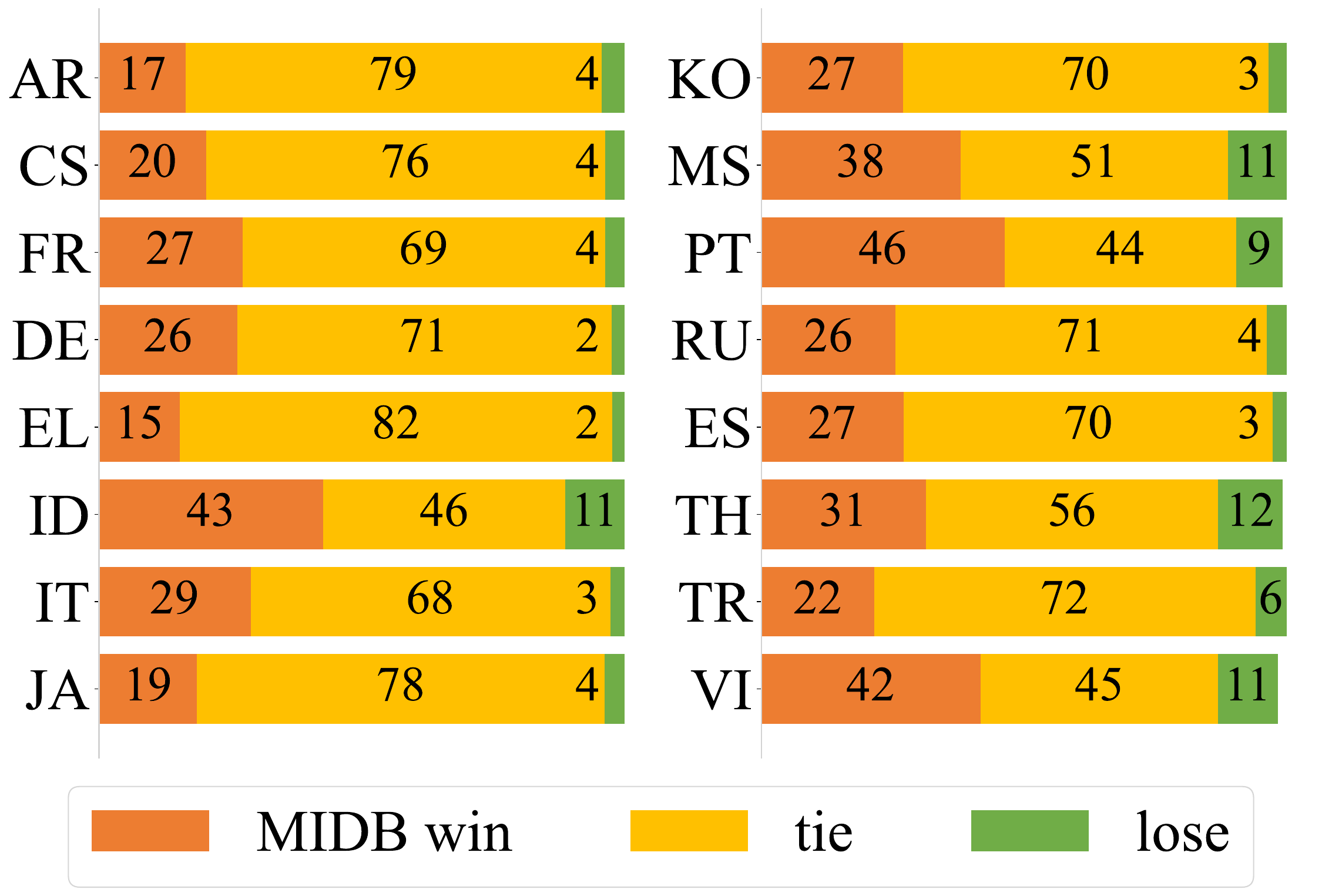}
    \caption{Win-lose-tie ratios of MIDB-boosted data compare with original data evaluated by LLM-as-judge.}
    \label{fig:data quality gpt}
\end{figure}

\subsection{Evaluation Approach}
\label{sec:Experiment Setting}

\paragraph{LLM-as-judge}
Following recent studies~\cite{ge2024clustering,liu2024coachlm,chen2023alpagasus}, we use advanced LLMs to automatically compare and score the responses of two candidates, based on the criteria from \citet{zheng2023judging} (detailed prompt in Appendix \ref{app: Prompt for GPT-4o as a Judge}). This prompt asks the LLM to judge the helpfulness, relevance, accuracy, and detail of each response (which is better) and provide a rationale. However, this automatic method suffers from reported evaluation biases when the order of candidates is changed~\cite{wang2023pandalm}. To alleviate the bias caused by positional deviation, we perform two evaluations swapping orders for each pair of candidates, and define the final judgment of a candidate as win only when it win twice, or win once and tie once (a win and a lose is counted as tie).

\paragraph{Human}

Seven experts independently rate each instruction pair or model response based on the criteria in Table~\ref{tab:MEDB_criteria}, unaware of the sources of the evaluated samples. They assess the satisfaction level across predefined dimensions and give a comparison-based judgment on two candidate samples. However, human evaluation is inherently limited in terms of efficiency and scalability, due to its high cost and the need for domain expertise. Further more, due to limited resources of senior language experts, we are unable to conduct human evaluation for a larger set of languages. Therefore, our language selection for evaluation covers both high-usage and low-resource languages for coverage and representativeness.

\paragraph{Metrics}
Several metrics are utilized for presenting the judgments: (1) By default we use \textit{win-lose-tie ratio}, which is given by $\frac{\#win}{\#all}$,$\frac{\#lose}{\#all}$ and $\frac{\#tie}{\#all}$, where $\#all$ is the number of test set samples; (2) For easy comparison between multiple baselines, we use \textit{win rate}, formulated as $\frac{\#win + \#tie}{\#all}$; and (3) \textit{winning score}, formulated as $\frac{\#win-\#lose}{\#all}+1$, to conveniently spot the winning side (score $>$ 1).

\subsection{Automatic Evaluation}\label{sec:GPT Evaluation}

\paragraph{Data Quality of MIDB-boosted Dataset}
\label{sec:Data Quality of MIDB-boosted Dataset}

We randomly sampled 520 data samples (strictly excluding those in MEB dataset) for each language from the machine-translated Alpaca dataset for quality assessment. An advanced LLM is employed to judge the winner for each original and MIDB-boosted data pair. Fig.~\ref{fig:data quality gpt} presents the win-lose-tie ratios of MIDB-boosted datasets in different languages. The results show that after MIDB boosting, all languages show a trend of significantly higher win ratios than lose ratios in terms of data quality. For example, Portuguese (PT) has a much higher winning ratio (46\%) than its losing ratio (9\%). This improvement suggests that the MIDB-boosted dataset primarily contains high-quality instruction pairs, which can enhance LLM instruction tuning while maintaining the original dataset's integrity.

\paragraph{Evaluation of LLMs Tuned on MIDB-boosted Dataset}
\label{sec:Evaluation of LLM Tuned on MIDB-boosted Dataset}

Starting from the full machine-translated Alpaca datasets (with 52k samples) in 16 languages, MIDB boosted each sample and resulted in MIDB-boosted Alpaca datasets for each language. Then, we trained two groups of models: (1) Alpaca-Original models~\cite{alpaca}, trained on LLaMA-3-8B~\cite{grattafiori2024llama3herdmodels} using the translated Alpaca datasets; (2) Alpaca-MIDB models, which are trained following the same setting as the Alpaca-Original models, but with the MIDB-boosted datasets replacing the translated Alpaca datasets. Several popular open-source LLMs that focus on multilingual tasks are also involved in the comparison. BayLing~\cite{zhang2023baylingbridgingcrosslingualalignment} and PolyLM~\cite{wei2023polylmopensourcepolyglot} are two multilingual LLMs with strong cross-lingual and multilingual performance on 10+ languages. DeepSeek-Multilingual~\cite{lightblue2023deepseek} is a recent model distilled from DeepSeek with extra training on multilingual data. 

Following Section~\ref{sec:construction_of_testset}, we assess them on two datasets extended by our experts: the AlpacaEval-16L and MT-Bench-16L, covering evaluations on both multilingual instruction following and multi-turn dialogues. All models are compared with reference answers generated by LLaMA3.1-8B-Instruct, a strong multilingual baseline through extensive training.

As shown in Fig.~\ref{fig:model ability}, Alpaca-MIDB performs exceptionally well on both rounds of the MT-Bench-16L benchmark, achieving notably high scores in low-resource languages such as Thai and Greek. On AlpacaEval-16L, Alpaca-MIDB also maintains its leading position averagely. However, due to the varying proficiency of the foundation model (\emph{i.e.}, LLaMA3-8B) across different languages, the performance of MIDB may exhibit fluctuations across certain language settings. Overall, Alpaca-MIDB exhibits strong multilingual instruction-following and multi-turn interactive capabilities, which originates from the high quality of its training data.

\subsection{Human Evaluation}~\label{sec:Human Evaluation}
In addition to automatic evaluation, seven human experts (their profiles described in Appendix~\ref{sec:detail_expert_profile}) independently assessed the quality of data and the performance of subsequently tuned Alpaca-MIDB.
\begin{figure}[htbp]
    \centering
    \includegraphics[width=0.9\linewidth]{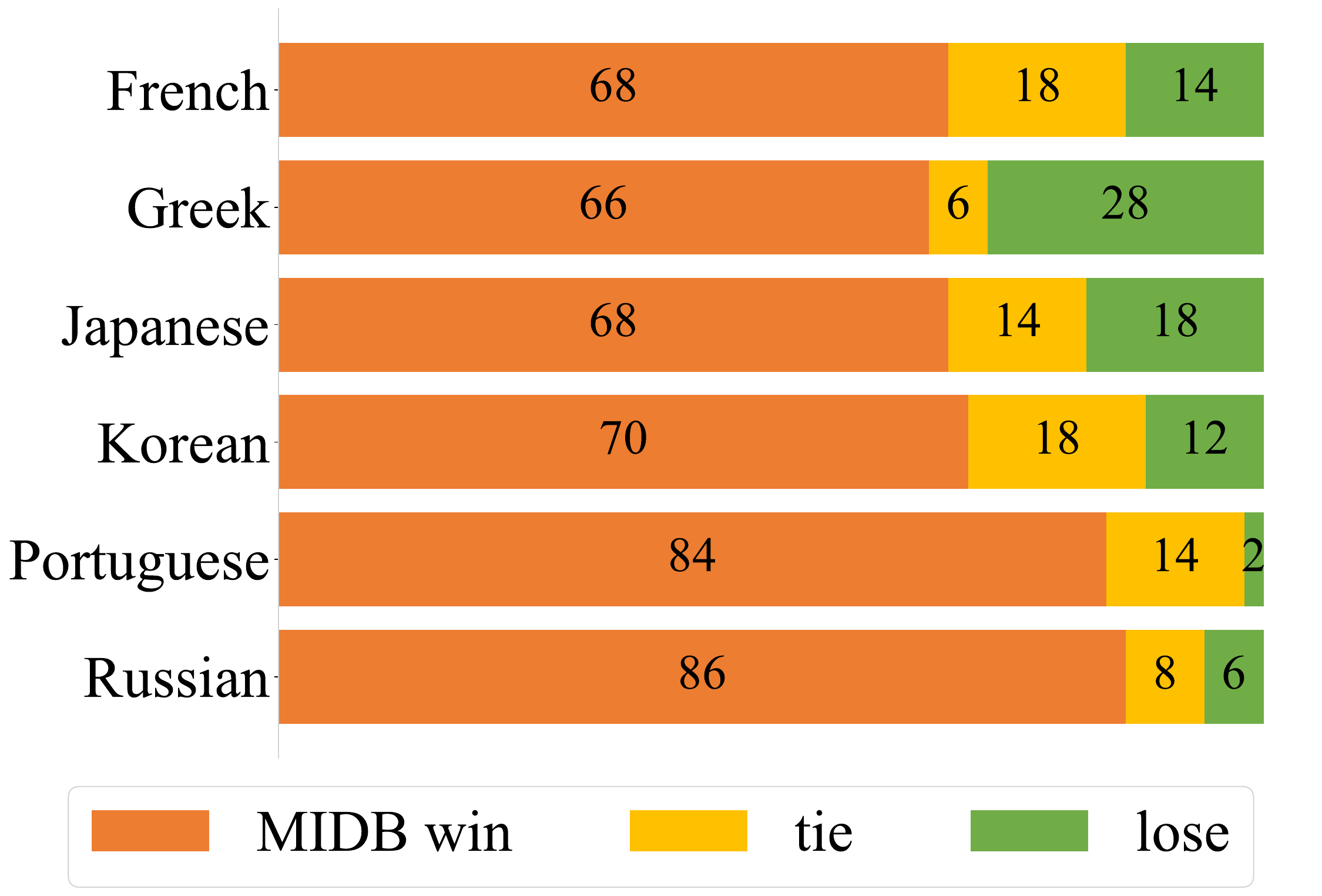}
    \caption{Win-lose-tie ratios of MIDB-boosted data compare with original data evaluated by human.}
    \label{fig:data quality human}
\end{figure}
 \paragraph{Data Quality} We randomly selected 50 instruction pairs from the original Alpaca datasets with their MIDB-boosted counterparts, and obtained independent ratings from multilingual experts who were unaware of the source of the samples. The results in Fig.~\ref{fig:data quality human} indicate that, after applying MIDB boosting, the data achieved a significantly higher average win ratios across all reviewers.

\begin{table}[htbp]
\centering
\resizebox{0.86\linewidth}{!}{
\begin{tabular}{lccc}
\toprule
\multirow{2}{*}{Language} & \multirow{2}{*}{AlpacaEval} & \multicolumn{2}{c}{MT-Bench} \\
\cmidrule(lr){3-4}
                           &                            & Round 1   & Round 2 \\
\midrule
French      & 1.68 & 1.56 & 1.52 \\
Greek       & 1.56 & 1.46 & 1.28 \\
Japanese    & 1.72 & 1.68 & 1.40 \\
Korean      & 1.36 & 1.38 & 1.38 \\
Portuguese  & 1.62 & 1.84 & 1.88 \\
Russian     & 1.68 & 1.72 & 1.52 \\
\bottomrule
\end{tabular}
}
\caption{Winning scores by human on models trained with MIDB-boosted data \emph{v.s.} original (score $>$ 1 is win).}
\label{tab: win rate human}
\end{table}

\paragraph{Alpaca-MIDB \emph{v.s.} Alpaca-Original} Human experts independently assessed the responses generated by Alpaca-MIDB and the original Alpaca-Original models on the AlpacaEval-16L and MT-Bench-16L test sets. The reviewers were unaware of the sources of the responses. As shown in Table~\ref{tab: win rate human}, human reviewers consistently gave Alpaca-MIDB higher ratings (winning scores ranging from 1.28 to 1.72) compared with the original Alpaca-Original model. This strong multilingual performance of Alpaca-MIDB further confirms the effectiveness of the boosts made by MIDB, leading to improved user experience in multilingual contexts.

\paragraph{Analysis of Human Reviews} It is worth noting that the enhancement brought by MIDB is consistently more pronounced across all 16 languages rated by human, as compared to ratings by LLM-as-judge. Thus, we conducted analysis on comments of reviewers and attributed it to nuanced but important improvements more perceivable to human, such as humanized tones and culturally appropriate expressions, which often leads to ties in LLM's judgments. Specifically, reviewers observed that responses of Alpaca-MIDB provided more detailed, human-like, well-structured, and readable content. Notably, Alpaca-MIDB was commented to have a richer reasoning process, particularly in addressing programming-related problems, resulting in superior outcomes quality.

\subsection{Evaluation of Alpaca-MIDB on Cultural Understanding Ability}~\label{sec: culture evaluation}
English-centric LLMs often suffer from limited localization capabilities from other cultures, as the most of instruction pairs used during training reflect the contexts of English, leading to linguistic and cultural bias when applied in multilingual environments. To verify the effectiveness of MIDB on enhancing cultural understanding, we employed BLEnD~\cite{myung2024blend}, a manually constructed question-answering benchmark specifically designed to evaluate LLMs' understanding of daily knowledge across diverse cultures and languages. The accuracy is calculated based on average performance under two prompts:
1) Directly ask the LLM to provide answers;
2) Add a role setting for the LLM to answer as a native in target culture.
We use an advanced LLM to verify if the LLM's response matches the manually labeled entity for each question (Appendix~\ref{app: Prompt for BLEnD Entity Detection}), and calculate the percentage of correct answers as the score. 
\begin{table}[htbp]
\centering
\resizebox{0.9\linewidth}{!}{
\begin{tabular}{lccc}
\toprule
Language & {Original} & {MIDB-Boosted} & {Up~$\uparrow$} \\
\midrule
Arabic      & 15.03 & 16.85 & 12.1\% \\
Greek       & 18.72 & 22.03 & 17.7\% \\
Spanish     & 25.00 & 28.49 & 14.0\% \\
Indonesian  & 20.62 & 25.30 & 22.7\% \\
Korean      & 18.50 & 24.19 & 30.8\% \\
\bottomrule
\end{tabular}
}
\caption{Accuracy score (0-100) of Alpaca-Original and Alpaca-MIDB on cultural-specificity knowledge.}
\label{tab:BLEnD result}
\end{table}

As shown in Table~\ref{tab:BLEnD result}, models trained on datasets boosted with MIDB achieve performance improvements ranging from 12.1\% to 30.8\%. These results highlight the effectiveness of MIDB in enriching training data with culturally relevant and localized knowledge, thereby significantly enhancing the models’ ability to understand and generate content that is contextually appropriate for non-English speakers.

\begin{figure}[htbp]
    \centering
    \includegraphics[width=.9\linewidth]{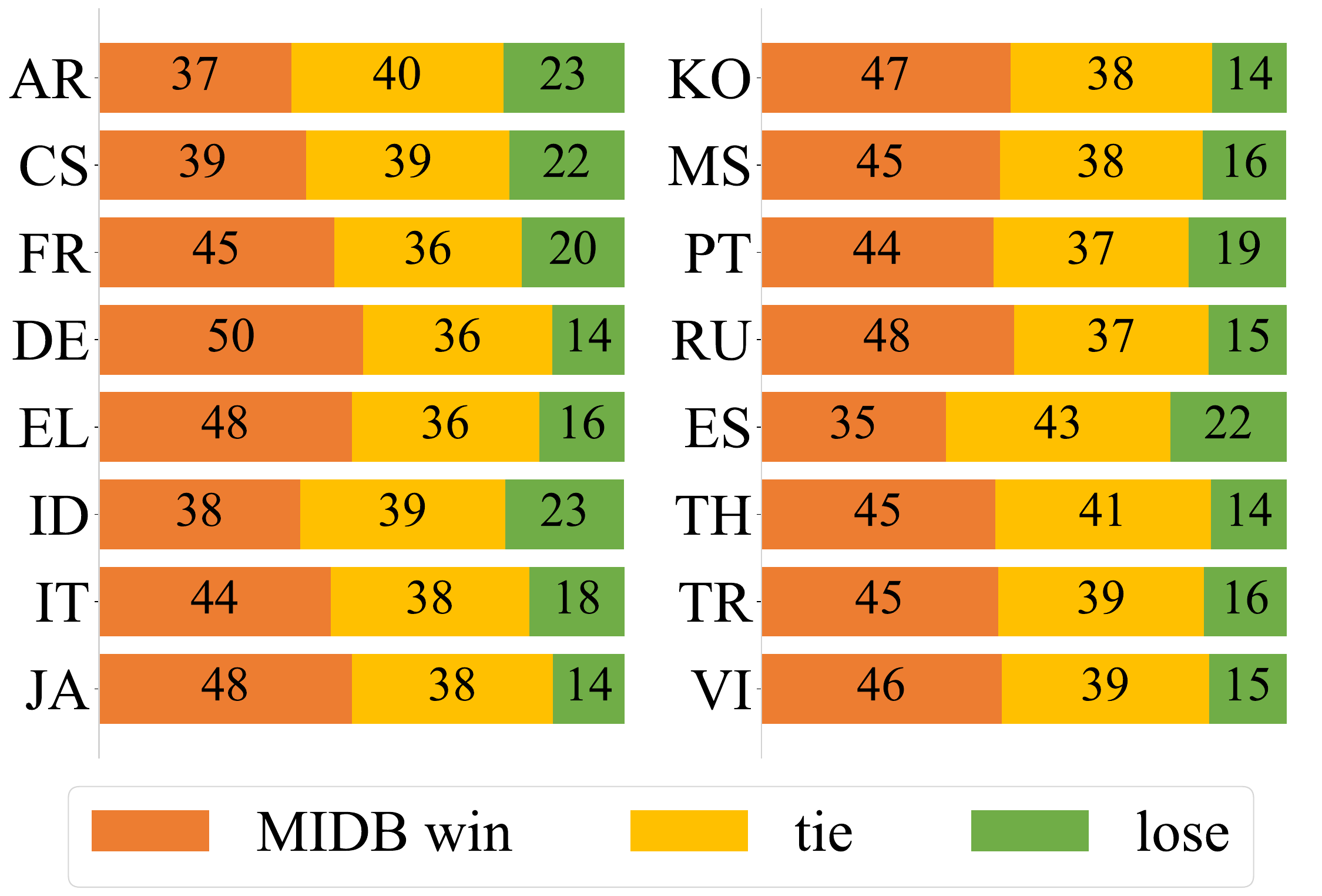}
    \caption{Win-lose-tie ratios of MIDB-boosted \emph{v.s.} original data on Dolly-15k, an out-of-distribution dataset.}
    \label{fig:OOD result}
\end{figure}
   
\subsection{Testing on Out-of-distribution Dataset}
\label{sec:Effectiveness on out-of-distribution dataset}
In addition to test MIDB on Alpaca-52k (in-distribution), we also select Dolly-15k~\cite{DatabricksBlog2023DollyV2}, a dataset composed of real-world user instructions on complex tasks such as brainstorming and creative writing to evaluate the generalization capability of our method on out-of-distribution (OOD) data. Compared with Alpaca-52k, which is synthesized using LLMs, Dolly-15k possesses a different data distribution by collecting samples from human. In Fig.~\ref{fig:OOD result}, despite OOD, the improvement of data quality after MIDB boosting remains significant, suggesting its strong generalization capabilities and promising application potentials.

\subsection{Sensitivity Analysis}\label{sec:albation_exp}

\begin{figure}[tbp]
 \centering  
 \subfigbottomskip=-2pt 
 \subfigcapskip=-2pt 
 \subfigure[Sensitivity on Varying Backbone Models $\theta$]{
  \includegraphics[width=0.85\linewidth]{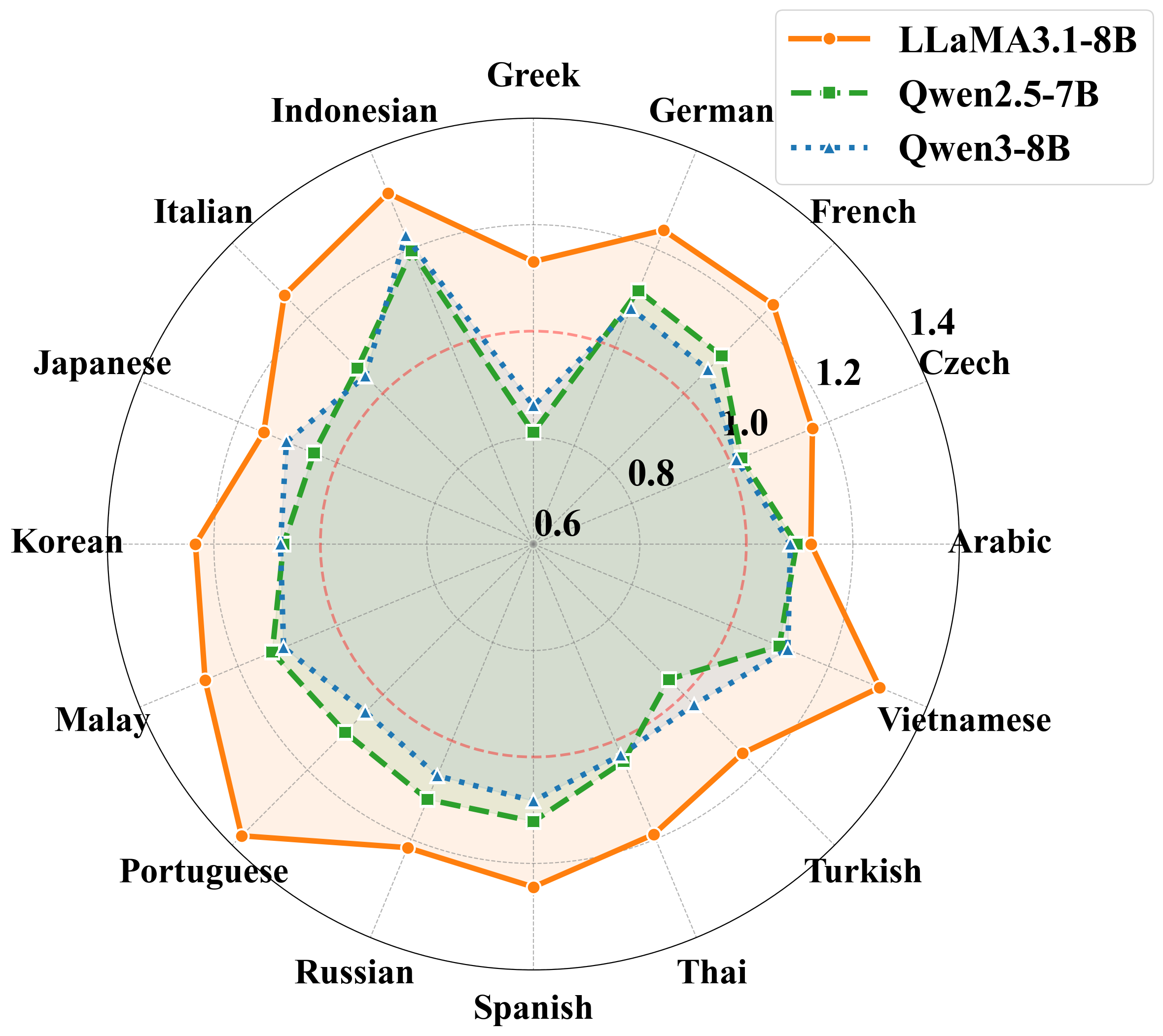}}
   \\
 \subfigure[Sensitivity on Quality-control Coefficient $\alpha$]{
  \includegraphics[width=0.85\linewidth]{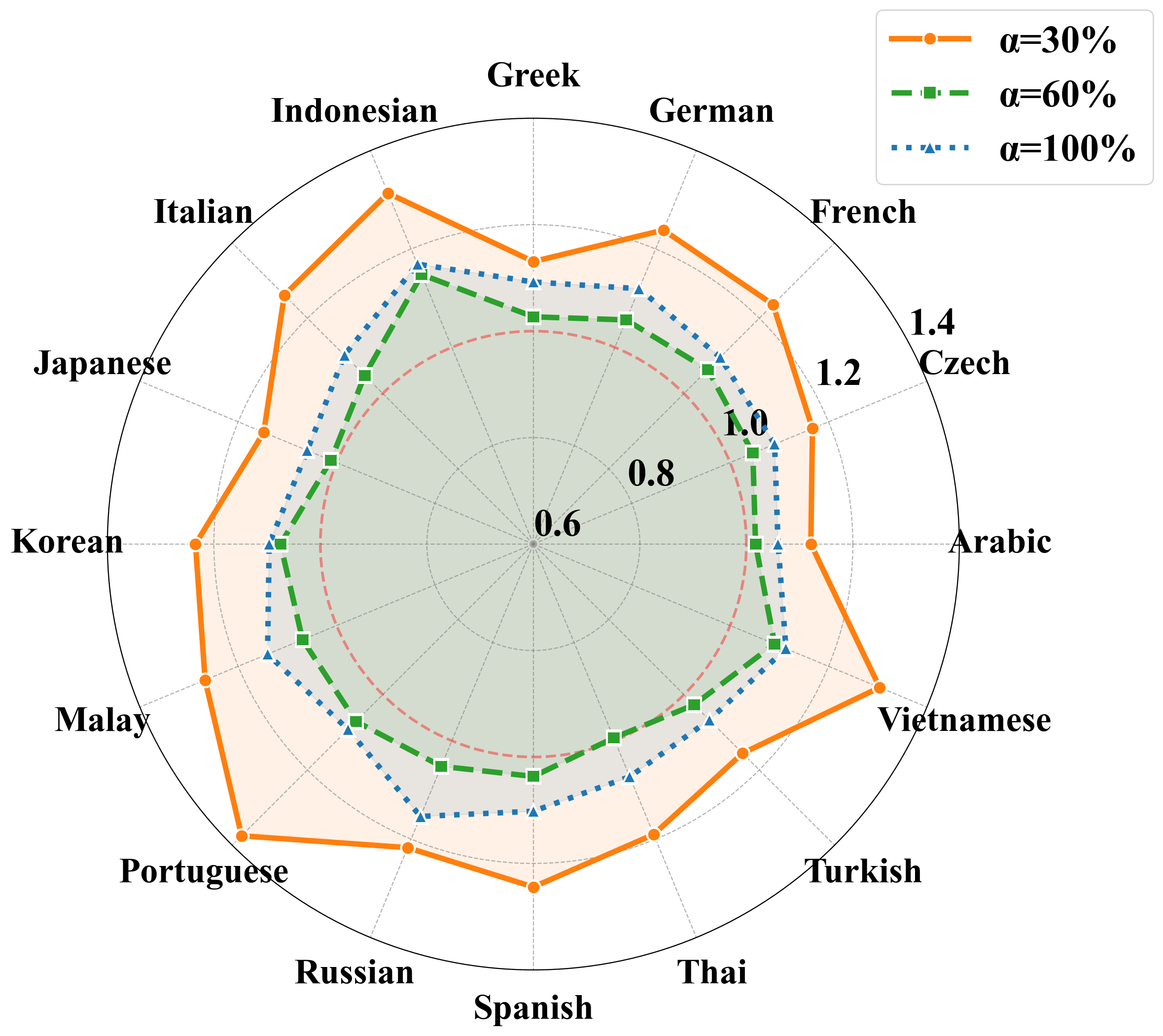}}
 \caption{Performance of MIDB with (a) varying backbone models and (b) varying quality-control ratios, displayed as winning score of MIDB-boosted data against pre-boosted data randomly selected from Alpaca dataset. Score $>$ 1 represents winning of MIDB (the dashed circle in red color).}
\label{fig_ablation}
\end{figure}

\paragraph{Variation on Backbone Model $\theta$}
To investigate the impact of different backbone models $\theta$ on the performance of MIDB, we conduct a comparison training MIDB using different backbone models: LLaMA3.1-8B-Instruct and two additional models from Qwen families, \emph{i.e.}, Qwen2.5-7B-Instruct~\cite{Yang2024Qwen25TR} and Qwen3-8B-Instruct~\cite{yang2025qwen3}. The evaluation setting (\emph{e.g.}, data sampling) is the same as Section~\ref{sec:Data Quality of MIDB-boosted Dataset}. As shown in Fig.~\ref{fig_ablation}(a), the winning scores, when Qwen family models serve as the backbone, exceeds one in most languages, indicating a valid data enhancement brought by MIDB with varying backbones. In addition, due to its strong multilingual capabilities, LLaMA3.1-8B shows a significant advantages against Qwen models. Based on these findings, we select LLaMA3.1-8B-Instruct as the base model of MIDB for our main experiments.

\paragraph{Variation on Quality-control Coefficient $\alpha$}
As discussed in Section~\ref{sec:training goal}, to ensure optimized performance of MIDB, we further incorporated the distance-based filtering mechanism in \citet{liu2024coachlm} when curating training data for MIDB, which is to retain only high-quality samples with the largest edit distances between the expert-revised instruction pairs and the original counterparts. The coefficient $\alpha$ controls the percentage of this high-quality subset. Following \citet{liu2024coachlm}, we directly adopt $\alpha$ as 30\%, which means that only the top 30\% samples with the most revisions will be used as for training MIDB. Fig.~\ref{fig_ablation}(b) displays the performances of MIDB with varying $\alpha$, using the evaluation setting in Section~\ref{sec:Data Quality of MIDB-boosted Dataset}. Despite varying $\alpha$, MIDB continuously achieves a winning score above one against original across languages, indicating a steady improvement on data quality after MIDB boosting. However, introducing more samples with less revision may lead to suboptimal performance of MIDB, as indicated by the relative advantages of “$\alpha$=30\%” group. This aligns with the hypothesis in \cite{liu2024coachlm} that manual revision samples with a higher edit distance contain more enriched learning patterns, while samples with a lower edit distance typically involve revisions limited to superficial aspects, such as grammar and layout adjustments.

\section{Conclusion}
In this paper, we proposed MIDB, an automatic tool to enhance linguistic and cultural equality in multilingual LLMs by addressing severe data quality issues such as cultural mismatching in synthesized instruction dataset. Experiment on 16 languages indicates universal improvement on data quality and subsequent model performance, suggesting a potential of MIDB to bridge the digital divide caused by English-centric AI technologies. Furthermore, the improvement on cultural understanding ability in the BLEnD test set renders MIDB a unique tool to mitigate cultural inequality concerns in high-stake areas such as education. Future work include expanding supported languages, testing on real-world scenarios and larger models.

\clearpage

\bibliography{MIDB}

\begin{thebibliography}{38}
\providecommand{\natexlab}[1]{#1}

\bibitem[{Achiam et~al.(2023)Achiam, Adler, Agarwal, Ahmad, Akkaya, Aleman, Almeida, Altenschmidt, Altman, Anadkat et~al.}]{achiam2023gpt}
Achiam, J.; Adler, S.; Agarwal, S.; Ahmad, L.; Akkaya, I.; Aleman, F.~L.; Almeida, D.; Altenschmidt, J.; Altman, S.; Anadkat, S.; et~al. 2023.
\newblock Gpt-4 technical report.
\newblock \emph{arXiv preprint arXiv:2303.08774}.

\bibitem[{Chen et~al.(2024{\natexlab{a}})Chen, Li, Yan, Wang, Gunaratna, Yadav, Tang, Srinivasan, Zhou, Huang, and Jin}]{chen2023alpagasus}
Chen, L.; Li, S.; Yan, J.; Wang, H.; Gunaratna, K.; Yadav, V.; Tang, Z.; Srinivasan, V.; Zhou, T.; Huang, H.; and Jin, H. 2024{\natexlab{a}}.
\newblock Alpagasus: Training a Better Alpaca Model with Fewer Data.
\newblock In \emph{International Conference on Learning Representations}.

\bibitem[{Chen et~al.(2024{\natexlab{b}})Chen, Ji, Bogoychev, Kutuzov, Haddow, and Heafield}]{chen2024monolingual}
Chen, P.; Ji, S.; Bogoychev, N.; Kutuzov, A.; Haddow, B.; and Heafield, K. 2024{\natexlab{b}}.
\newblock Monolingual or Multilingual Instruction Tuning: Which Makes a Better Alpaca.
\newblock In \emph{Findings of the Association for Computational Linguistics: EACL 2024}, 1347--1356.

\bibitem[{Chen et~al.(2024{\natexlab{c}})Chen, Yu, Guo, and Haddow}]{chen2024good}
Chen, P.; Yu, S.; Guo, Z.; and Haddow, B. 2024{\natexlab{c}}.
\newblock Is It Good Data for Multilingual Instruction Tuning or Just Bad Multilingual Evaluation for Large Language Models?
\newblock In \emph{Proceedings of the 2024 Conference on Empirical Methods in Natural Language Processing}, 9706--9726.

\bibitem[{Conover et~al.(2023)Conover, Hayes, Mathur, Xie, Wan, Shah, Ghodsi, Wendell, Zaharia, and Xin}]{DatabricksBlog2023DollyV2}
Conover, M.; Hayes, M.; Mathur, A.; Xie, J.; Wan, J.; Shah, S.; Ghodsi, A.; Wendell, P.; Zaharia, M.; and Xin, R. 2023.
\newblock Free Dolly: Introducing the World's First Truly Open Instruction-Tuned LLM.

\bibitem[{DeepSeek-AI(2025)}]{deepseek2025r1}
DeepSeek-AI. 2025.
\newblock DeepSeek-R1: Incentivizing Reasoning Capability in LLMs via Reinforcement Learning.
\newblock In \emph{arXiv preprint arXiv:2501.12948}.

\bibitem[{Dubois et~al.(2023)Dubois, Li, Taori, Zhang, Gulrajani, Ba, Guestrin, Liang, and Hashimoto}]{dubois2023alpacafarm}
Dubois, Y.; Li, C.~X.; Taori, R.; Zhang, T.; Gulrajani, I.; Ba, J.; Guestrin, C.; Liang, P.~S.; and Hashimoto, T.~B. 2023.
\newblock Alpacafarm: A simulation framework for methods that learn from human feedback.
\newblock \emph{Advances in Neural Information Processing Systems}, 36: 30039--30069.

\bibitem[{Feng et~al.(2025)Feng, Gao, Chen, Chen, and Shang}]{feng2025culfit}
Feng, R.; Gao, S.; Chen, X.; Chen, L.; and Shang, S. 2025.
\newblock CulFiT: A Fine-grained Cultural-aware LLM Training Paradigm via Multilingual Critique Data Synthesis.
\newblock \emph{arXiv preprint arXiv:2505.19484}.

\bibitem[{Ge et~al.(2024)Ge, Liu, Hu, Meng, Tao, Zhao, Xia, Li, Chen, Yang et~al.}]{ge2024clustering}
Ge, Y.; Liu, Y.; Hu, C.; Meng, W.; Tao, S.; Zhao, X.; Xia, M.; Li, Z.; Chen, B.; Yang, H.; et~al. 2024.
\newblock Clustering and Ranking: Diversity-preserved Instruction Selection through Expert-aligned Quality Estimation.
\newblock In \emph{Proceedings of the 2024 Conference on Empirical Methods in Natural Language Processing}, 464--478.

\bibitem[{Grattafiori et~al.(2024)Grattafiori, Dubey, Jauhri, Pandey, Kadian, Al-Dahle, Letman, Mathur, Schelten, Vaughan et~al.}]{grattafiori2024llama3herdmodels}
Grattafiori, A.; Dubey, A.; Jauhri, A.; Pandey, A.; Kadian, A.; Al-Dahle, A.; Letman, A.; Mathur, A.; Schelten, A.; Vaughan, A.; et~al. 2024.
\newblock The Llama 3 Herd of Models.
\newblock arXiv:2407.21783.

\bibitem[{Hu et~al.(2021)Hu, Shen, Wallis, Allen-Zhu, Li, Wang, Wang, and Chen}]{hu2021loralowrankadaptationlarge}
Hu, E.~J.; Shen, Y.; Wallis, P.; Allen-Zhu, Z.; Li, Y.; Wang, S.; Wang, L.; and Chen, W. 2021.
\newblock LoRA: Low-Rank Adaptation of Large Language Models.
\newblock arXiv:2106.09685.

\bibitem[{Huo et~al.(2025)Huo, Feng, Huang, Fu, Li, Ye, Zhang, Tu, Tang, Lu et~al.}]{huo2025enhancing}
Huo, W.; Feng, X.; Huang, Y.; Fu, C.; Li, B.; Ye, Y.; Zhang, Z.; Tu, D.; Tang, D.; Lu, Y.; et~al. 2025.
\newblock Enhancing Non-English Capabilities of English-Centric Large Language Models through Deep Supervision Fine-Tuning.
\newblock \emph{arXiv preprint arXiv:2503.01275}.

\bibitem[{Joshi et~al.(2020)Joshi, Santy, Budhiraja, Bali, and Choudhury}]{joshi2020state}
Joshi, P.; Santy, S.; Budhiraja, A.; Bali, K.; and Choudhury, M. 2020.
\newblock The State and Fate of Linguistic Diversity and Inclusion in the NLP World.
\newblock In \emph{Proceedings of the 58th Annual Meeting of the Association for Computational Linguistics}, 6282--6293.

\bibitem[{Lai, Mesgar, and Fraser(2024)}]{lai2024llms}
Lai, W.; Mesgar, M.; and Fraser, A. 2024.
\newblock LLMs beyond English: Scaling the multilingual capability of LLMs with cross-lingual feedback.
\newblock \emph{arXiv preprint arXiv:2406.01771}.

\bibitem[{Li et~al.(2024)Li, Zhang, Li, Chen, Chen, Cheng, Wang, Zhou, and Xiao}]{li2024quantity}
Li, M.; Zhang, Y.; Li, Z.; Chen, J.; Chen, L.; Cheng, N.; Wang, J.; Zhou, T.; and Xiao, J. 2024.
\newblock From Quantity to Quality: Boosting LLM Performance with Self-Guided Data Selection for Instruction Tuning.
\newblock In \emph{Proceedings of the 2024 Conference of the North American Chapter of the Association for Computational Linguistics: Human Language Technologies (Volume 1: Long Papers)}, 7595--7628.

\bibitem[{Li et~al.(2023)Li, Zhang, Dubois, Taori, Gulrajani, Guestrin, Liang, and Hashimoto}]{alpaca_eval}
Li, X.; Zhang, T.; Dubois, Y.; Taori, R.; Gulrajani, I.; Guestrin, C.; Liang, P.; and Hashimoto, T.~B. 2023.
\newblock AlpacaEval: An Automatic Evaluator of Instruction-following Models.
\newblock \url{https://github.com/tatsu-lab/alpaca_eval}.

\bibitem[{LightBlue(2025)}]{lightblue2023deepseek}
LightBlue. 2025.
\newblock DeepSeek-R1-Distill-Qwen-7B-Multilingual.

\bibitem[{Liu et~al.(2024)Liu, Tao, Zhao, Zhu, Ma, Zhu, Su, Hou, Zhang, Zhang et~al.}]{liu2024coachlm}
Liu, Y.; Tao, S.; Zhao, X.; Zhu, M.; Ma, W.; Zhu, J.; Su, C.; Hou, Y.; Zhang, M.; Zhang, M.; et~al. 2024.
\newblock Coachlm: Automatic instruction revisions improve the data quality in llm instruction tuning.
\newblock In \emph{2024 IEEE 40th International Conference on Data Engineering (ICDE)}, 5184--5197. IEEE.

\bibitem[{Lynch(2025)}]{stanford2025digitaldivide}
Lynch, S. 2025.
\newblock How AI is leaving non-English speakers behind.
\newblock \emph{Stanford News}.

\bibitem[{Myung et~al.(2024)Myung, Lee, Zhou, Jin, Putri, Antypas, Borkakoty, Kim, Perez-Almendros, Ayele et~al.}]{myung2024blend}
Myung, J.; Lee, N.; Zhou, Y.; Jin, J.; Putri, R.; Antypas, D.; Borkakoty, H.; Kim, E.; Perez-Almendros, C.; Ayele, A.~A.; et~al. 2024.
\newblock Blend: A benchmark for llms on everyday knowledge in diverse cultures and languages.
\newblock \emph{Advances in Neural Information Processing Systems}, 37: 78104--78146.

\bibitem[{Ruebsamen(2023)}]{AlpacaDataCleaned}
Ruebsamen, G. 2023.
\newblock Cleaned Alpaca Dataset.
\newblock GitHub repository.

\bibitem[{Rystr{\o}m, Kirk, and Hale(2025)}]{rystrom2025multilingual}
Rystr{\o}m, J.; Kirk, H.~R.; and Hale, S. 2025.
\newblock Multilingual!= multicultural: Evaluating gaps between multilingual capabilities and cultural alignment in llms.
\newblock \emph{arXiv preprint arXiv:2502.16534}.

\bibitem[{Shaham et~al.(2024)Shaham, Herzig, Aharoni, Szpektor, Tsarfaty, and Eyal}]{shaham2024multilingual}
Shaham, U.; Herzig, J.; Aharoni, R.; Szpektor, I.; Tsarfaty, R.; and Eyal, M. 2024.
\newblock Multilingual Instruction Tuning With Just a Pinch of Multilinguality.
\newblock In \emph{Findings of the Association for Computational Linguistics ACL 2024}, 2304--2317.

\bibitem[{Taori et~al.(2023)Taori, Gulrajani, Zhang, Dubois, Li, Guestrin, Liang, and Hashimoto}]{alpaca}
Taori, R.; Gulrajani, I.; Zhang, T.; Dubois, Y.; Li, X.; Guestrin, C.; Liang, P.; and Hashimoto, T.~B. 2023.
\newblock Stanford Alpaca: An Instruction-following LLaMA model.
\newblock https://github.com/tatsu-lab/stanford\_alpaca.

\bibitem[{Touvron et~al.(2023{\natexlab{a}})Touvron, Lavril, Izacard, Martinet, Lachaux, Lacroix, Rozi{\`e}re, Goyal, Hambro, Azhar et~al.}]{touvron2023llama}
Touvron, H.; Lavril, T.; Izacard, G.; Martinet, X.; Lachaux, M.-A.; Lacroix, T.; Rozi{\`e}re, B.; Goyal, N.; Hambro, E.; Azhar, F.; et~al. 2023{\natexlab{a}}.
\newblock Llama: Open and efficient foundation language models.
\newblock \emph{arXiv preprint arXiv:2302.13971}.

\bibitem[{Touvron et~al.(2023{\natexlab{b}})Touvron, Martin, Stone, Albert, Almahairi, Babaei, Bashlykov, Batra, Bhargava, Bhosale et~al.}]{touvron2023llama2}
Touvron, H.; Martin, L.; Stone, K.; Albert, P.; Almahairi, A.; Babaei, Y.; Bashlykov, N.; Batra, S.; Bhargava, P.; Bhosale, S.; et~al. 2023{\natexlab{b}}.
\newblock Llama 2: Open foundation and fine-tuned chat models.
\newblock \emph{arXiv preprint arXiv:2307.09288}.

\bibitem[{{\"U}st{\"u}n et~al.(2024){\"U}st{\"u}n, Aryabumi, Yong, Ko, D’souza, Onilude, Bhandari, Singh, Ooi, Kayid et~al.}]{ustun2024aya}
{\"U}st{\"u}n, A.; Aryabumi, V.; Yong, Z.; Ko, W.-Y.; D’souza, D.; Onilude, G.; Bhandari, N.; Singh, S.; Ooi, H.-L.; Kayid, A.; et~al. 2024.
\newblock Aya Model: An Instruction Finetuned Open-Access Multilingual Language Model.
\newblock In \emph{Proceedings of the 62nd Annual Meeting of the Association for Computational Linguistics (Volume 1: Long Papers)}, 15894--15939.

\bibitem[{Wang et~al.(2023{\natexlab{a}})Wang, Kordi, Mishra, Liu, Smith, Khashabi, and Hajishirzi}]{wang-etal-2023-self-instruct}
Wang, Y.; Kordi, Y.; Mishra, S.; Liu, A.; Smith, N.~A.; Khashabi, D.; and Hajishirzi, H. 2023{\natexlab{a}}.
\newblock Self-Instruct: Aligning Language Models with Self-Generated Instructions.
\newblock In \emph{Proceedings of the 61st Annual Meeting of the Association for Computational Linguistics (Volume 1: Long Papers)}, 13484--13508. Toronto, Canada: Association for Computational Linguistics.

\bibitem[{Wang et~al.(2023{\natexlab{b}})Wang, Yu, Zeng, Yang, Wang, Chen, Jiang, Xie, Wang, Xie et~al.}]{wang2023pandalm}
Wang, Y.; Yu, Z.; Zeng, Z.; Yang, L.; Wang, C.; Chen, H.; Jiang, C.; Xie, R.; Wang, J.; Xie, X.; et~al. 2023{\natexlab{b}}.
\newblock Pandalm: An automatic evaluation benchmark for llm instruction tuning optimization.
\newblock \emph{arXiv preprint arXiv:2306.05087}.

\bibitem[{Wei et~al.(2023)Wei, Wei, Lin, Li, Zhang, Ren, Li, Wan, Cao, Xie, Hu, Li, Hui, Yu, Liu, Yang, Huang, and Xie}]{wei2023polylmopensourcepolyglot}
Wei, X.; Wei, H.; Lin, H.; Li, T.; Zhang, P.; Ren, X.; Li, M.; Wan, Y.; Cao, Z.; Xie, B.; Hu, T.; Li, S.; Hui, B.; Yu, B.; Liu, D.; Yang, B.; Huang, F.; and Xie, J. 2023.
\newblock PolyLM: An Open Source Polyglot Large Language Model.
\newblock arXiv:2307.06018.

\bibitem[{Xu et~al.(2024)Xu, Sun, Zheng, Geng, Zhao, Feng, Tao, Lin, and Jiang}]{xu2023wizardlm}
Xu, C.; Sun, Q.; Zheng, K.; Geng, X.; Zhao, P.; Feng, J.; Tao, C.; Lin, Q.; and Jiang, D. 2024.
\newblock Wizard{LM}: Empowering Large Pre-Trained Language Models to Follow Complex Instructions.
\newblock In \emph{International Conference on Learning Representations}.

\bibitem[{Yang et~al.(2025)Yang, Li, Yang, Zhang, Hui, Zheng, Yu, Gao, Huang, Lv et~al.}]{yang2025qwen3}
Yang, A.; Li, A.; Yang, B.; Zhang, B.; Hui, B.; Zheng, B.; Yu, B.; Gao, C.; Huang, C.; Lv, C.; et~al. 2025.
\newblock Qwen3 technical report.
\newblock \emph{arXiv preprint arXiv:2505.09388}.

\bibitem[{Yang et~al.(2024)Yang, Yang, Zhang, Hui, Zheng, Yu, Li, Liu, Huang, Dong, Wei, Lin, Yang, Tu, Zhang, Yang, Yang, Zhou, Lin, Dang, Lu, Bao, Yang, Yu, Li, Xue, Zhang, Zhu, Men, Lin, Li, Xia, Ren, Ren, Fan, Su, Zhang, Wan, Liu, Cui, Zhang, Qiu, Quan, and Wang}]{Yang2024Qwen25TR}
Yang, Q.~A.; Yang, B.; Zhang, B.; Hui, B.; Zheng, B.; Yu, B.; Li, C.; Liu, D.; Huang, F.; Dong, G.; Wei, H.; Lin, H.; Yang, J.; Tu, J.; Zhang, J.; Yang, J.; Yang, J.; Zhou, J.; Lin, J.; Dang, K.; Lu, K.; Bao, K.; Yang, K.; Yu, L.; Li, M.; Xue, M.; Zhang, P.; Zhu, Q.; Men, R.; Lin, R.; Li, T.; Xia, T.; Ren, X.; Ren, X.; Fan, Y.; Su, Y.; Zhang, Y.-C.; Wan, Y.; Liu, Y.; Cui, Z.; Zhang, Z.; Qiu, Z.; Quan, S.; and Wang, Z. 2024.
\newblock Qwen2.5 Technical Report.
\newblock \emph{ArXiv}, abs/2412.15115.

\bibitem[{Zhang et~al.(2023)Zhang, Fang, Zhang, Ma, Zhou, Huang, Bu, Gui, Chen, Chen, and Feng}]{zhang2023baylingbridgingcrosslingualalignment}
Zhang, S.; Fang, Q.; Zhang, Z.; Ma, Z.; Zhou, Y.; Huang, L.; Bu, M.; Gui, S.; Chen, Y.; Chen, X.; and Feng, Y. 2023.
\newblock BayLing: Bridging Cross-lingual Alignment and Instruction Following through Interactive Translation for Large Language Models.
\newblock arXiv:2306.10968.

\bibitem[{Zhang et~al.(2024)Zhang, Lee, Fang, Yu, Jia, Jiang, and Barbieri}]{zhang2024plug}
Zhang, Z.; Lee, D.-H.; Fang, Y.; Yu, W.; Jia, M.; Jiang, M.; and Barbieri, F. 2024.
\newblock PLUG: Leveraging Pivot Language in Cross-Lingual Instruction Tuning.
\newblock In \emph{Proceedings of the 62nd Annual Meeting of the Association for Computational Linguistics (Volume 1: Long Papers)}, 7025--7046.

\bibitem[{Zheng et~al.(2023)Zheng, Chiang, Sheng, Zhuang, Wu, Zhuang, Lin, Li, Li, Xing et~al.}]{zheng2023judging}
Zheng, L.; Chiang, W.-L.; Sheng, Y.; Zhuang, S.; Wu, Z.; Zhuang, Y.; Lin, Z.; Li, Z.; Li, D.; Xing, E.; et~al. 2023.
\newblock Judging llm-as-a-judge with mt-bench and chatbot arena.
\newblock \emph{Advances in Neural Information Processing Systems}, 36: 46595--46623.

\bibitem[{Zhou et~al.(2024)Zhou, Liu, Xu, Iyer, Sun, Mao, Ma, Efrat, Yu, Yu et~al.}]{zhou2023lima}
Zhou, C.; Liu, P.; Xu, P.; Iyer, S.; Sun, J.; Mao, Y.; Ma, X.; Efrat, A.; Yu, P.; Yu, L.; et~al. 2024.
\newblock Lima: Less is more for alignment.
\newblock \emph{Advances in Neural Information Processing Systems}, 36.

\bibitem[{Zhu et~al.(2023)Zhu, Lv, Dong, Yuan, Xu, Huang, Kong, Chen, and Li}]{zhu2023extrapolating}
Zhu, W.; Lv, Y.; Dong, Q.; Yuan, F.; Xu, J.; Huang, S.; Kong, L.; Chen, J.; and Li, L. 2023.
\newblock Extrapolating large language models to non-english by aligning languages.
\newblock \emph{arXiv preprint arXiv:2308.04948}.

\end{thebibliography}



\appendix

\section{Limitations}
\label{sec:lim}

Despite the strong performance of MIDB across 16 languages, including several low-resource ones, its current evaluation and design scope remain limited in several important respects:

\textbf{Limited Language Coverage.}
Currently, MIDB only supports 16 languages, taking up only a small number of languages in the world, which is a compromise due to our available expert resources. However, as shown in Appendix~\ref{sec:language_distribution}, our language selection strategies emphasize both geographical grouping and popularity levels, ensuring a comprehensive coverage within our budget. Given the promising results on the 16 languages, we plan to gradually build supports for more languages in a long-term base.

\textbf{Hard to Scale due to Human-labeling Cost.}
The human-labeling nature of MIDB's training set makes it difficult to scale to other languages. However, without human intervention, it is impractical to resolve the deeper data quality issues in synthesized dataset such as content hallucinations and cultural localization. Compared with full dataset annotation, MIDB only utilizes a small amount of human-labeling efforts (around 5\% of the Alpaca dataset), and yet achieves significantly improvements on the data quality and model performance. To further reduce the cost for language scaling, strategies such as global crowd-sourcing can be taken to ensure an efficient automatic data boosting in the target language with only small number of annotations by human.

\textbf{Insufficient Support for Advanced Reasoning and Computational Tasks.}
Current evaluations have been limited to instruction-following benchmarks on general topics. Consequently, MIDB’s capacity to handle more cognitively demanding tasks—such as multi-step logical reasoning, mathematical computations, or compositional decision-making—has yet to be systematically explored or validated. However, due to the significant differences between general instruction and reasoning-based tasks, major adaptions on training designs of MIDB are expected to be made in order to support complex reasoning tasks. We mark this as an interesting topic for future exploration. 

\section{Ethical Consideration}
\label{sec:consideration}

This work involves the curation and revision of multilingual instruction data across 16 languages. The following key aspects were considered:

\textbf{Ethical and Professional Annotation Practices.}
All human-annotated revisions used in training MIDB were carried out by qualified linguistic experts under fair working conditions with appropriate compensation. This ensured high-quality data while upholding ethical standards.

\textbf{Bias Awareness and Cultural Sensitivity.}
During the data correction and localization process care was taken to avoid introducing cultural biases or normative judgments. However residual biases may still persist. These could originate from the initial synthetic data or be introduced during the revision process.

\textbf{Challenges and Future Directions.}
While MIDB aims to improve linguistic and cultural alignment there is a risk that automated systems may reinforce dominant language norms or overlook underrepresented dialects. Future efforts should involve broader community participation especially from speakers of low-resource and marginalized languages. This is essential for promoting more inclusive and equitable multilingual AI development.

\section{Detailed Analysis on Profile of Involved Multilingual Human Experts and Task Allocation Strategy}\label{sec:detail_expert_profile}

\begin{table}[htbp]
\centering
\resizebox{\linewidth}{!} {
\begin{tabular}{p{0.6\linewidth} p{0.2\linewidth} p{0.2\linewidth} c}
\toprule
\textbf{Assigned Tasks} & Experts & \textbf{Years} & \textbf{Type} \\
 \midrule
Construction of MEB Dataset & 23 & 8.4 & Outsourced \\
 \midrule
Localization of Benchmarks & 20 & 5.2 & Outsourced \\
 \midrule
Human Evaluation & 7 & 3.9 & Regular \\
\bottomrule
\end{tabular}
}
\caption{Profiles and allocations of language experts}
\label{tab:group_summary}
\end{table}

\begin{table}[tbp]
\centering
\resizebox{\linewidth}{!}{
\begin{tabular}{lcll}
\toprule
Language & Code & Geographical Groupings               & Popularity Level   \\
\midrule
French & FR & Western Europe                       & High Resource      \\
German & DE & Central Europe                       & High Resource      \\
Czech & CS & Central Europe                        & Medium Resource       \\
Greek & EL & Southern Europe                       & Low Resource       \\
Italian & IT & Southern Europe                      & Medium Resource      \\
Portuguese & PT & Southern Europe                     & Medium Resource    \\
Spanish & ES & Southern Europe                      & High Resource      \\
Russian & RU & EENA     & Medium Resource      \\
Turkish & TR & EENA    & Medium Resource    \\
Arabic & AR & MENA  & High Resource    \\
Japanese & JA & East Asia                            & High Resource    \\
Korean & KO & East Asia                            & Medium Resource    \\
Indonesian & ID & Southeast Asia                      & Low Resource       \\
Malay & MS & Southeast Asia                        & Low Resource       \\
Thai & TH & Southeast Asia                         & Low Resource       \\
Vietnamese & VI & Southeast Asia                     & Medium Resource       \\
\bottomrule
\end{tabular}
}
\caption{Language names mapping to codes, geographical grouping and resource level, where MENA represents Middle East and North Africa; EENA represents Eastern Europe and Northern Asia.}
\label{tab:languages_table}
\end{table}
The expertise of our multilingual team was pivotal in ensuring the quality of the datasets and evaluations. As shown in Table~\ref{tab:group_summary}, we recruited a group of highly skilled language experts with diverse linguistic backgrounds from the language service center of a prominent international corporation. All experts are educated professionals in linguistics, offering services such as translation, localization, editing and technical writing for multiple languages.

The allocation of experts into the three tasks in Table~\ref{tab:group_summary} was carefully structured, considering both task complexity and experts' language proficiency. Among the tasks, human evaluations on dataset quality and model performance are expected to be the most challenging, requiring mastery of target language and consistent internal discussion, which renders only regular employees eligible for this task. To ensure a broad language coverage of MIDB, outsourced experts were also involved for the task of training set construction and benchmark localization, with the former allocated more resource due to relative task complexity. The expertise assigned to each language is ensured to be evenly distributed (\emph{i.e.}, possibly multiple experts assigned to the same language). Notably, there is a strict worker split between the three tasks to avoid potential bias in evaluations (\emph{e.g.}, if the same outsourced vendor revised training data went further localizing the benchmarks, there could be bias).

During this tasks, quality assurance was maintained under the rigorous procedure of the cooperated language service center. Specifically, a two-round review-rebuttal feedback loop was applied for each language, with third-party reviewers continuously inspecting samples during annotation and discussing with annotators to reach agreements.

\section{Supported Languages}\label{sec:appendix B}
\subsection{Language Name-Code Mappings}\label{language_dict}

Table~\ref{tab:languages_table} displays the mappings between full names of languages and their short codes, as well as geographical and resource-level information for each language.

\begin{figure}[tbp]
    \centering
    \includegraphics[width=0.85\linewidth]{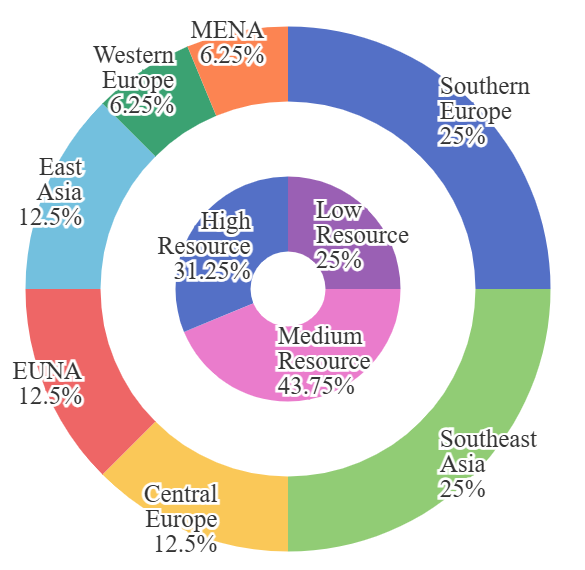}
    \caption{Languages distribution, where MENA represents Middle East and North Africa; EUNA represents Eastern Europe and Northern Asia.}
    \label{fig:languages_distribution}
\end{figure}
\subsection{Reasons for Language Selection}\label{sec:language_distribution}
As shown in Fig.~\ref{fig:languages_distribution}, The 16 languages represent regions across Europe, Asia and Africa, which are relatively underrepresented by existing AI technologies. Additionally, as shown in Table~\ref{tab:languages_table}, we include languages with lower popularity and fewer resources, such as Greek, Indonesian, Thai, and Malay. The classification of popularity level in Table~\ref{tab:languages_table} is according to the taxonomy built by~\citet{joshi2020state}, which rates 2485 languages based on its available linguistic resources. We label those who receive a 5-score as high-resource languages (\emph{e.g.}, Spanish), those with a 4-score as middle-resource languages (\emph{e.g.}, Korean), and those scored less than 3 as low-resource languages. As a result, our dataset provides a broader generalization across a diverse set of languages, thus enhancing both the languages coverage of the model and the credibility of evaluation process.

\section{Prompts}
\subsection{Prompt for LLM-as-Judge}
\label{app: Prompt for GPT-4o as a Judge}
\begin{mybox}
\verb|[System]| \\
Please act as an impartial judge and evaluate the quality of the responses provided by two AI assistants to the user question displayed below. You should choose the assistant that follows the user’s instructions and answers the user’s question better. Your evaluation should consider factors such as the helpfulness, relevance, accuracy, depth, creativity, and level of detail of their responses. Begin your evaluation by comparing the two responses and provide a short explanation. Avoid any position biases and ensure that the order in which the responses were presented does not influence your decision. Do not allow the length of the responses to influence your evaluation. Do not favor certain names of the assistants. Be as objective as possible. After providing your explanation, output your final verdict by strictly following this format: "[[1]]" if assistant A is better, "[[2]]" if assistant B is better, and "[[0]]" for a tie. \\
\verb|[User Question]| \\
{\{question\}} \\
\verb|[The Start of Assistant A’s Answer]| \\
{\{answer\_a\}} \\
\verb|[The End of Assistant A’s Answer]| \\
\verb|[The Start of Assistant B’s Answer]| \\
{\{answer\_b\}} \\
\verb|[The End of Assistant B’s Answer]|
\end{mybox}

\subsection{Prompt for BLEnD Entity Detection}
\label{app: Prompt for BLEnD Entity Detection}
\begin{mybox}
\verb|[System]| \\
Check whether an entity in the specified set is matched in a given statement. Return "Yes" or "No" only in the first line of the answer.\\

\verb|[Given statement]| \\
{\{model\_answer\}}

\verb|[Given collection]| \\
{\{answe\_set\}}
\end{mybox}

\end{document}